\documentclass{article}

\usepackage[final]{corl_2026} 

\newcommand{\ourmethod}{SidewalkBench\xspace}
\newcommand{\urbansim}{Urban-Sim\xspace}
\newcommand{\metaurban}{MetaUrban\xspace}

\usepackage{booktabs}
\usepackage{multirow}
\usepackage{graphicx}
\usepackage{amsmath}
\usepackage{xspace}
\usepackage{wrapfig}
\usepackage[table]{xcolor}
\usepackage{color, colortbl}
\usepackage{lipsum} 
\usepackage{caption}
\usepackage{xcolor}

\usepackage{titlesec}
\usepackage{enumitem}
\titlespacing*{\paragraph}
{0pt}{0.3ex}{0.7em}

\title{\ourmethod: Benchmarking Visual Navigation on Urban Sidewalks}

\author{
Zhizheng Liu$^{1,}$\thanks{Equal contribution.} \qquad
Honglin He$^{1, *}$ \qquad
Vivek Alumootil$^{1, *}$ \\
\textbf{Akshat Pandya}$^{2}$ \qquad
\textbf{Brad Squicciarini}$^{2}$ \qquad
\textbf{Wayne Wu}$^{1}$ \qquad
\textbf{Bolei Zhou}$^{1,2}$ \\
$^{1}$University of California, Los Angeles \qquad $^{2}$Coco Robotics\\
{\url{https://vail.cs.ucla.edu/SidewalkBench/}} \\
}

\begin{document}
\maketitle
\vspace{-8mm}

\begin{abstract}
Urban sidewalk navigation presents significant challenges due to complex structural layouts, dynamic pedestrian behaviors, and long distances. While recent visual navigation models offer a promising solution, the lack of a unified benchmark hinders quantitative and reproducible evaluation. To bridge this gap, we propose \ourmethod, a comprehensive benchmark designed for visual navigation on urban sidewalks. Built upon NVIDIA Isaac Sim, \ourmethod brings GPU-accelerated simulation of diverse, high-fidelity sidewalk environments, including both procedurally generated and real-world scanned scenes. We further populate the scenes with rich, reactive event-based pedestrian behaviors and flexible, efficient animation, enabling standardized model evaluation under realistic real-world settings.
%
We conduct a comprehensive evaluation of 9 visual navigation models on 330 unit-test scenarios, 800 pedestrian-reactive scenarios, and 105 long-horizon scenarios.  
Our findings highlight that pedestrian interaction and long-horizon robustness remain critical bottlenecks for existing models, and scaling up sidewalk training with synthetic data emerges as a promising solution.
     
\end{abstract}
 \vspace{-2mm}

\vspace{-4mm}

\begin{figure}[h]
    \centering
    \includegraphics[width=1.0\textwidth]{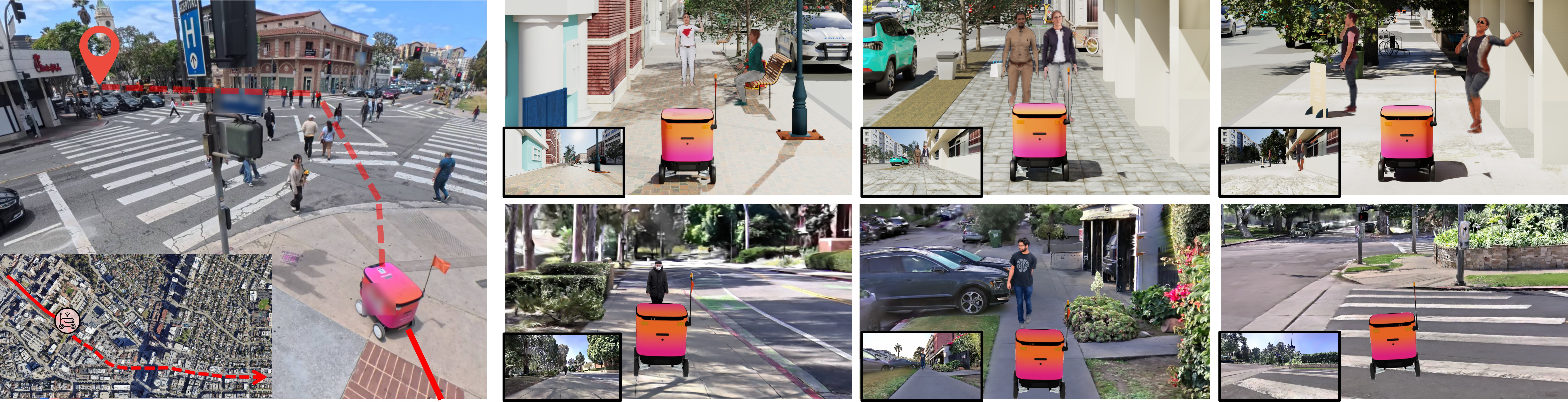}
    \caption{Left: Challenges of sidewalk navigation include static obstacles, dynamic pedestrians, diverse structures and layouts, and long-horizon distances. Right: \ourmethod can simulate diverse and realistic scenarios for standardized evaluation of visual navigation on urban sidewalks. }
    \label{fig:teaser}
 \vspace{-4mm}
\end{figure}
 
\section{Introduction}
\vspace{-2mm}

%

The prevalence of mobile robots and devices on urban sidewalks is rapidly increasing~\cite{abduljabbar2021role}. For instance, sidewalk delivery robots streamline last-mile logistics, electric wheelchairs provide accessible transportation to the elderly and people with disabilities, and e-scooters offer eco-friendly commuting alternatives. Transitioning to the autonomous operation of these machines can further reduce labor costs and improve personal convenience. Yet, the ability to safely navigate complex city streets still remains a significant challenge:
As shown in Fig.~\ref{fig:teaser}~(left), the robot must navigate long distances through several street blocks with varied structures and layouts, cross intersections while obeying traffic rules and signs, and avoid collisions with obstacles such as lamp posts and stop signs. 
More importantly, sidewalks are populated with pedestrians who exhibit complex behaviors and movements~\cite{daamen2003experimental}, introducing a high degree of uncertainty that requires the robot to predict pedestrians' intent and interact safely and socially.

Recent visual navigation foundation models~\cite{shah2023vint, hirose2025learning, he2025seeing, wei2026ground} have provided a promising solution for urban sidewalk navigation, requiring only a monocular RGB camera for perception and demonstrating strong generalization across environments and embodiments.
Although a few visual navigation methods have been evaluated on urban sidewalks separately~\cite{he2025seeing, liu2025citywalker, payandeh2025narrate2nav, huang2026tic}, the scale of their experiments is rather limited, and it is difficult to assess the model's capability for real-world deployment. 
Specifically, their testing scenarios are usually short-horizon trials and lack structural and layout diversity. 
For instance, NavBench-GS~\cite{he2025seeing} only includes straight paths mostly less than 10 meters.
Furthermore, there are no standardized protocols and explicit scenario definitions essential for reproducible, comprehensive analysis.
Crucially, no current study has isolated how different pedestrian behaviors would affect the model performance. For example,  CityWalker~\cite{liu2025citywalker} simply classifies the scenarios into forward, left turn, and right turn.
While existing social navigation benchmarks~\cite{pirk2022protocol, tsoi2022sean, perez2023hunavsim} do analyze human-robot interaction across various social scenarios, they primarily focus on indoor settings and are not dedicated to urban sidewalks.
Consequently, the lack of a unified sidewalk navigation benchmark has become a critical gap for fairly comparing models and isolating the failure modes and bottlenecks that guide the future development of autonomous robots in public spaces.

In this work, we aim to establish a comprehensive benchmark for visual navigation on urban sidewalks: \ourmethod. We build our benchmark on NVIDIA Isaac Sim~\cite{NVIDIA_Isaac_Sim}, taking full advantage of its accurate physics simulation, efficient parallel rendering, and extensive simulation ecosystem such as UrbanVerse~\cite{liu2026urbanverse} and \urbansim~\cite{wu2025urbansim}. To simulate diverse and realistic urban scenes, \ourmethod introduces two distinct types of environments: (1)  procedurally generated urban scenes that compose various sidewalk blocks, layouts, and obstacles into diverse large-scale environments, and (2)  real-world scanned scenes reconstructed from 3D Gaussian Splatting (3DGS)~\cite{kerbl20233d} via a high-fidelity spatial camera to ensure high visual and geometric realism. To generate rich pedestrian behaviors and movements, we further develop a pedestrian simulation module that includes a set of dynamic, event-based behaviors triggered to generate diverse pedestrian-interactive scenarios, as well as a custom pedestrian animation pipeline with high flexibility and efficiency. Ultimately, as shown in Fig.~\ref{fig:teaser}~(right), \ourmethod enables the simulation of diverse scenes and pedestrian activities, reflecting the challenges of real-world sidewalk navigation.

To comprehensively evaluate different aspects of model performance, \ourmethod introduces three distinct types of testing scenarios: 330 unit-test scenarios, 800 pedestrian-reactive scenarios, and 105 long-horizon scenarios. We benchmark 9 representative visual navigation models across these scenarios and conduct an in-depth analysis. Our main findings reveal that: (1) scaling urban sidewalk training data is the most important factor for the visual navigation performance; (2) nearly all models struggle with complex pedestrian behaviors and gesture understanding; (3) long-horizon navigation remains far from solved, with the top-performing model still experiencing 0.68  failures per 100m of travel distance; and (4) our simulation platform can serve as a scalable synthetic data generator for model finetuning.  These insights highlight a critical gap for future research, particularly in enhancing pedestrian behavior understanding and long-horizon robustness.


We summarize our main contributions as follows:
\begin{itemize}[nosep, topsep=0pt]
    \item A comprehensive benchmark \ourmethod for visual navigation on urban sidewalks, facilitating standardized model testing and comparison.
    \item An urban sidewalk simulation platform with diverse and realistic scenes as well as rich and efficient pedestrian simulation, reflecting the challenges in real-world deployment.
    \item  A detailed evaluation and analysis of representative visual navigation models on unit-test scenarios, pedestrian-reactive scenarios, and long-horizon scenarios, revealing the lack of pedestrian behavior understanding and long-horizon robustness in existing models.
\end{itemize}

\vspace{-2mm}

\section{Related Work}
\vspace{-2mm}

\paragraph{Visual Navigation Foundation Models}
Visual navigation foundation models~\cite{shah2023vint,he2025seeing,wei2026ground,liu2025citywalker,shah2023gnm,sridhar2024nomad,he2026learning,wei2025streamvln} use inputs from a monocular RGB camera and predict control signals or waypoints for robot navigation.
These models are usually trained end-to-end using a data-driven approach, with many recent advances in model architecture and training recipes. NoMaD~\cite{sridhar2024nomad} and NavDP~\cite{cai2025navdp} introduce diffusion policies for navigation and exploration. CityWalker~\cite{liu2025citywalker} and NWM~\cite{bar2025navigation} use web-scale video data for broad generalization. S2E~\cite{he2025seeing} introduces reinforcement learning with a simulator for post-training, and recent Vision-Language-Action (VLA) models~\cite{wei2026ground,payandeh2025narrate2nav,huang2026tic,agarwal2026cosmos} have emerged to map continuous visual streams and linguistic instructions directly to real-time continuous control signals. 
Despite promising progress, there is no uniform benchmark for evaluating visual navigation in urban sidewalk scenarios, which is an important downstream application for these models.



\paragraph{Simulation Benchmarks for Visual Navigation}
Simulation benchmarks for visual navigation enable standardized, reproducible comparisons of different models. Existing benchmarks~\cite{deitke2022, habitatchallenge2023}  mainly focus on static indoor scenes, and
follow-up works~\cite{gong2025cognition, li2024human} incorporate dynamic human behaviors to evaluate human-aware navigation performance. To bridge the Real2Sim evaluation gap with higher visual realism,  Vid2Sim~\cite{xie2025vid2sim}, NavBench-GS~\cite{he2025seeing}, and Wanderland~\cite{liu2025wanderland} use reconstructed outdoor scenarios from 3DGS~\cite{kerbl20233d} to provide real-world grounded simulation but lack dynamic pedestrians.
SocNavBench~\cite{biswas2022socnavbench} includes scenarios with urban pedestrians, but it suffers from low rendering quality and limited scene diversity. 
Furthermore, these benchmarks are typically limited to short-horizon evaluations with trajectories spanning around 10 meters~\cite{he2025seeing,biswas2022socnavbench}. Our \ourmethod is the first of its kind  that focuses specifically on large-scale urban dynamic environments with both procedurally generated and real-world scanned scenes.

\paragraph{Urban Simulators for Embodied AI}
With the growing interest in developing autonomous agents in urban public spaces, many urban simulators have been developed with a focus on various tasks and embodiments including autonomous driving~\cite{dosovitskiy2017carla}, UAVs~\cite{ge2025airsim360}, sidewalk autonomy~\cite{wu2025urbansim,wu2025metaurban}, and LLM/VLM agents~\cite{gao2024embodied, zhuang2026simworld}. 
Some social navigation simulators~\cite{tsoi2022sean, kastnerarena, escudero2025hunavsim} are also relevant which focus on simulating various pedestrian behaviors.
For the evaluation of visual navigation on urban sidewalks, the simulation engine is required to simulate accurate physics and support efficient and realistic rendering of the RGB observations, while many of the simulators above don't support these functionalities using game engines like Unity~\cite{unity2026} or Unreal~\cite{unreal2026}.
Built upon Isaac Sim~\cite{NVIDIA_Isaac_Sim} and Isaac Lab~\cite{mittal2025isaaclab}, our \ourmethod leverages the high-fidelity, GPU-accelerated physics engines tailored for robot learning to simulate large-scale urban scenes and provide realistic RGB observations in parallel environments. To address the pedestrian rendering bottleneck, we further develop a more flexible and efficient pedestrian animation pipeline compared to prior works~\cite{wu2025urbansim, escudero2025hunavsim}.


\vspace{-2mm}

\section{\ourmethod Design}
\vspace{-2mm}

\ourmethod uses NVIDIA Isaac Sim~\cite{NVIDIA_Isaac_Sim} as the simulation engine, which provides GPU acceleration to enable accurate physics simulation and realistic camera rendering at scale. To generate diverse and realistic urban sidewalk testing scenarios, we first introduce our sidewalk scenes in Sec.~\ref{sec:sidewalk_scenario}. Within these scenes, we further simulate rich pedestrian behaviors with diverse human-robot interactions, which is detailed in Sec.~\ref{sec:ped_simulation}. Based on the simulation platform, we define a diverse set of scenarios in Sec.~\ref{sec:tasks} that the models might often encounter in real-world deployment.


\begin{figure}[t]
    \centering
    \includegraphics[width=1.0\textwidth]{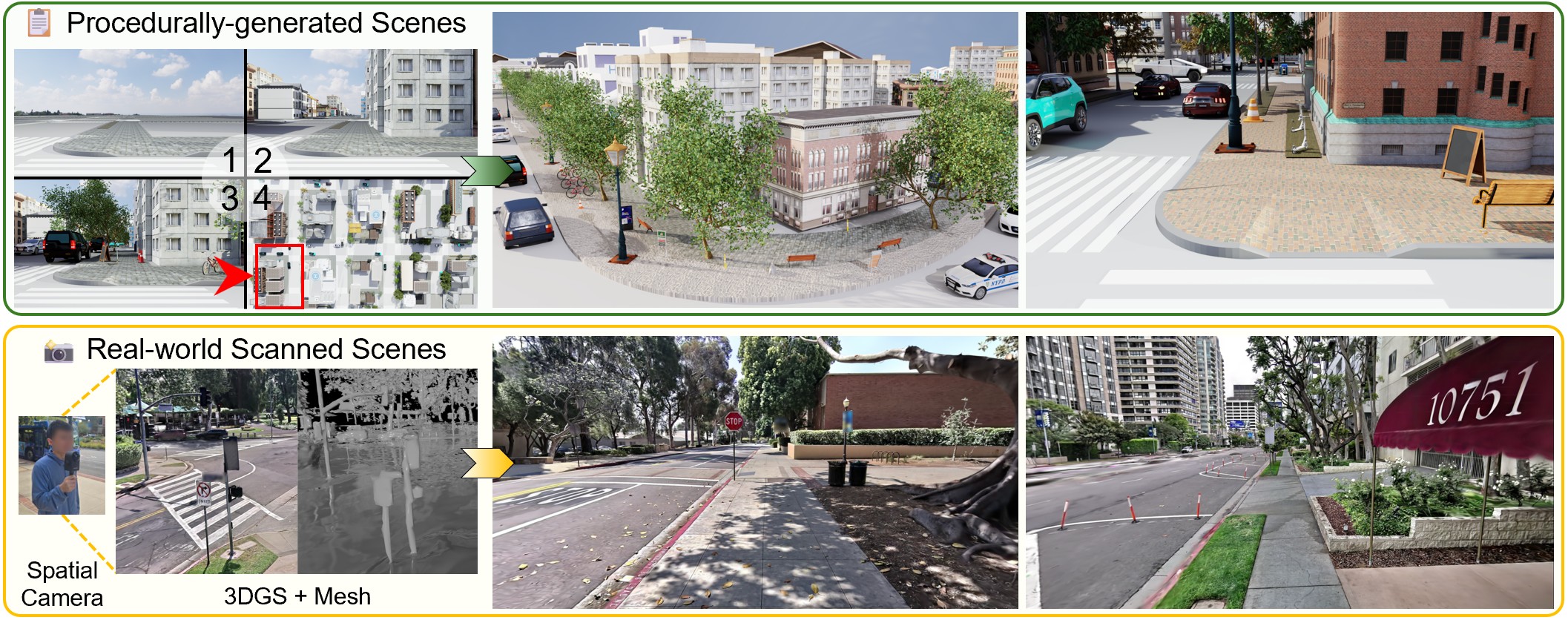}
    \vspace{-4mm}
    \caption{ \textbf{Overview of   scene types in \ourmethod.} }
        \vspace{-4mm}
    \label{fig:platform}
\end{figure}
\vspace{-2mm}

\subsection{Scene Types}
\vspace{-2mm}

\ourmethod includes two complementary scene types: procedurally generated scenes and real-world scanned scenes. The procedurally generated scenes focus on the diversity and controllability of sidewalk structures and layouts, enabling a comprehensive evaluation of the model's generalization performance. The real-world scanned scenes prioritize the realism of scene appearance and geometry, allowing comparisons under photorealistic conditions and facilitating the study of the real-to-sim gap.  An overview of the scenes is illustrated in Fig.~\ref{fig:platform}.

\label{sec:sidewalk_scenario}

\paragraph{Procedurally Generated Scenes}
\label{sec:scene_generation}
We use procedural generation to create large-scale scenes with diverse sidewalk blocks, layouts, and obstacles. Following \metaurban~\cite{wu2025metaurban} and \urbansim~\cite{wu2025urbansim}, we first define 7 primitive block types, such as straight segments, curves, and intersections with varying lengths, then connect them via spline-based routing to form continuous urban topologies. Next, we split each block into 5 functional zones, including roads, sidewalks, curbs and gutters, road verges, and frontage zones, according to the lateral functional structure of the block. We further randomize the layouts of these zones and add block-specific elements, such as ramps and crosswalks at intersections. Finally, we leverage UrbanVerse-100K~\cite{liu2026urbanverse}, a large-scale urban asset database, to sample diverse sky HDRIs, ground textures, physical materials, and zone-specific static objects with randomized placements. Using this pipeline, we generate 100 large-scale environments, each covering an area of 2~km$\times$2~km and comprising a wide variety of sidewalk configurations.



\paragraph{Real-world Scanned Scenes} Recent advancements in 3DGS~\cite{kerbl20233d} have enabled robot training and evaluation~\cite{xie2025vid2sim,escontrela2025gaussgym} with photo-realistic rendering. To further reconstruct accurate geometry, we use a spatial camera from XGRIDS~\cite{xgrids_portalcam}, equipped with a LiDAR and four cameras, to scan and reconstruct many street blocks. The reconstructed scenes feature realistic visual appearance from 3DGS and accurate sidewalk geometry and physics from the scanned mesh. After that, we annotate the sidewalk and crosswalk regions and convert the reconstruction to a simulation-ready format supported by NVIDIA Isaac Sim~\cite{NVIDIA_Isaac_Sim}. In total, we collected 11 real-world scanned scenes with various block types and the average scale is  150~m$\times$150~m.
\vspace{-2mm}
\subsection{Pedestrian Simulation}
\vspace{-2mm}
\label{sec:ped_simulation}

Pedestrian simulation is crucial for testing a model's capability to safely react to the diverse pedestrian behaviors and motions encountered in real-world urban spaces. 
\ourmethod adopts a two-level approach for pedestrian simulation:
For high-level behaviors, we introduce event-based behaviors to generate standardized human-interactive scenarios with diverse behaviors.
For low-level animation, we propose a new pipeline with superior movement flexibility and rendering efficiency.
More details on the pedestrian simulation can be found in Appendix~\ref{supp:nav_scenario}.

\begin{figure}[t]
    \centering
    \includegraphics[width=1.0\textwidth]{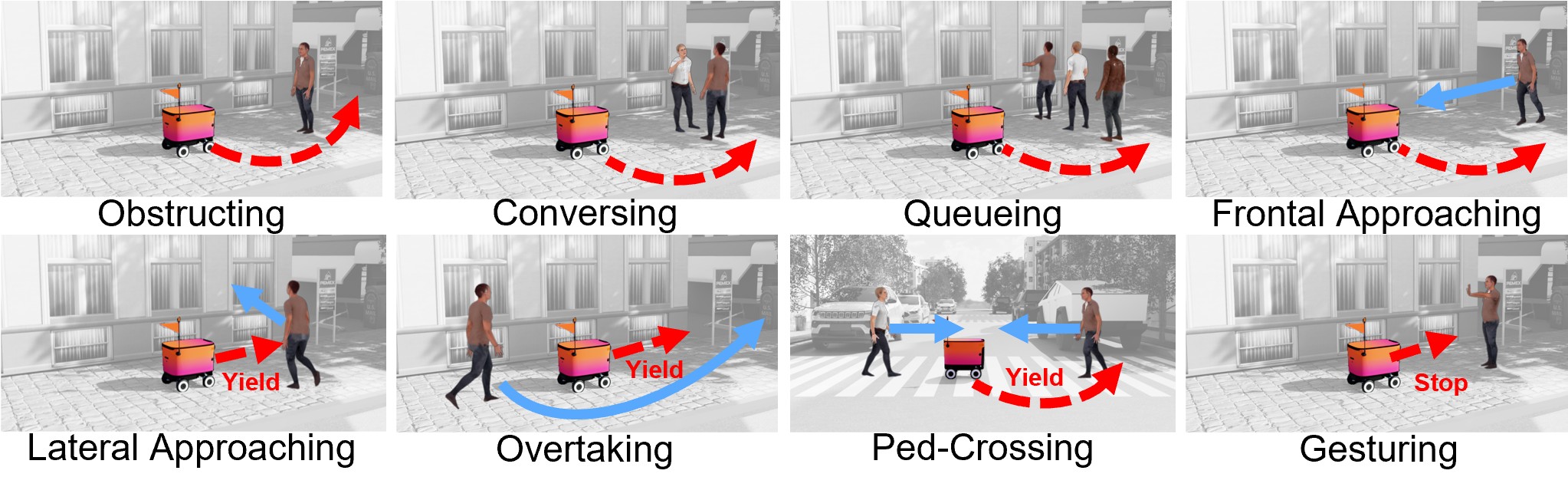}
    \vspace{-6mm}
    \caption{\textbf{Illustration of the event-based behaviors}. The blue arrow indicates the moving direction of the pedestrian after the event is triggered and the red arrow indicates the robot's intended reaction.}
        \vspace{-6mm}

    \label{fig:benchmark}
\end{figure}

\paragraph{Event-based High-level Behaviors}   We use behavior graphs~\cite{tsoi2022sean} to model the high-level pedestrian behavior, where each node represents either a navigation waypoint or a static object that a pedestrian can interact with. A state machine determines the behavior transitions between walking between nodes, staying idle, or interacting with the object or other pedestrians.  
%
%
To further enrich pedestrian behaviors and create human-interactive scenarios for the robot, we propose event-based pedestrian behaviors that are dynamically triggered by the pedestrian's relative position to the robot.  
This allows us to test the robot's reaction under different pedestrian behaviors and generate standardized scenarios for reproducible benchmarking.
%
%
Specifically, we classify common interaction behaviors on urban sidewalks, including both existing social scenarios defined in~\cite{pirk2022protocol, francis2025principles} and sidewalk-specific behaviors such as pedestrian crossing and queueing, each with its own triggering condition. 
We then incorporate these event-based behaviors into the behavior state machine to simulate a highly diverse range of behaviors.
An illustration of the behaviors is shown in Fig.~\ref{fig:benchmark}.

\paragraph{Flexible and Efficient Low-level Animation}  Pedestrian movements in existing urban simulators~\cite{wu2025urbansim,wu2025metaurban,zhuang2026simworld} primarily rely on fixed animation assets with limited movement types such as walking and standing, which lack the diversity and flexibility to generate arbitrary movements such as the stopping gesture. 
 In addition, the pedestrian rendering speed has become a critical bottleneck for simulation efficiency with the complex graphics pipeline.
 \ourmethod designs a new pedestrian animation pipeline in which we represent all pedestrians using the SMPL~\cite{loper2023smpl} human body model and directly use the SMPL parameters to control pedestrian animation and render either textured SMPL meshes~\cite{casas2023smplitex} or SMPL-based human 3DGS avatars~\cite{kocabas2024hugs, jiang2022neuman}.  This greatly enhances the flexibility and diversity of pedestrian movements, with full control over detailed motion, by leveraging human motion generation models~\cite{dai2024motionlcm} and large-scale human motion datasets~\cite{lin2023motion, liu2025learning}. We further develop a custom pedestrian renderer based on the highly efficient Nvdiffrast library~\cite{Laine2020diffrast} and compose the pedestrian rendering with the background rendered by the simulator, achieving a 60x improvement in rendering efficiency compared to the character behavior module~\cite{nvidia_isaac_sim_actor_control} in Isaac Sim, thereby enabling large-scale evaluation in parallel environments. 

\vspace{-2mm}
\subsection{Scenarios}
\vspace{-2mm}
\label{sec:tasks}
Current visual navigation models are typically tested on a few short-horizon sidewalk scenarios with limited scene diversity~\cite{he2025seeing, liu2025citywalker}, which can hardly reflect real-world deployment performance. Moreover, it is unclear how these models would react to various pedestrian behaviors without a standardized, reproducible benchmark. 
Based on the simulation platform discussed above, \ourmethod proposes three complementary scenario types for comprehensive model evaluation: unit-test scenarios, pedestrian-reactive scenarios, and long-horizon scenarios. It also allows users to define custom scenarios to evaluate specific sidewalk navigation tasks. More details are presented in Appendix~\ref{supp:nav_scenario}.
%

\textbf{Unit-test scenarios} assess the model's performance in navigating basic topological structures of sidewalks. These scenarios are short-horizon with the distances between the start and the goal ranging from 10 to 20 meters and are sampled from three types of sidewalk blocks: straight blocks, curve blocks, and crosswalks. We focus on evaluating the models' lane-following and static collision-avoidance capabilities in different structures and layouts, and therefore do not simulate pedestrians in these scenarios. The task fails if the robot collides with any obstacles or is outside the sidewalk region, and succeeds if the robot reaches the goal within the time limit (60s). In total, we collect 330 unit-test scenarios with 100 scenarios for each block type in the procedurally generated scenes and 10 scenarios for each block type in the real-world scanned scenes.

\textbf{Pedestrian-reactive scenarios}  evaluate the model's ability to safely react to the various types of pedestrian behaviors. The setting is similar to the unit-test navigation, where we additionally simulate one or more pedestrians with different event-based behaviors (Fig.~\ref{fig:benchmark}) and place them on the path to the goal to ensure the event can be triggered. We only evaluate on the procedurally generated scenes, as they are more structured and controllable for standardized testing.  Except for the pedestrian crossing behavior, which is evaluated at the crosswalk, all other behaviors are tested on the straight sidewalk block to isolate the effects of structural variations. In total, we collect 800 pedestrian-reactive scenarios with 100 scenarios for each type of pedestrian behavior.

\textbf{Long-horizon scenarios} require the robot to traverse large-scale environments to mimic real-world deployment. We sample the start and goal at a long distance ($>$100m) and ensure there is only a single path from the start to the goal by blocking the other routes with obstacles. As this is a challenging task for current visual navigation models, we do not terminate the episode upon failure; instead, we reset the robot to the closest waypoint and count the failure modes. In total, we collect 105 long-horizon scenarios with 100 scenarios from procedurally generated scenes and 5 scenarios from real-world scanned scenes—covering a total route distance of 36.5 km.


\paragraph{Evaluation Metrics} For unit-test scenarios, we follow existing navigation benchmarks to evaluate route completion rate~\cite{habitatchallenge2023, liu2025wanderland,dauner2024navsim}. 
For pedestrian-reactive scenarios, we evaluate the success rate for overall task completion. For long-horizon scenarios, similar to how human intervention frequency is used to evaluate real-world autonomous driving~\cite{kohanpour2025trends}, we evaluate the average failure counts per 100 meters traveled, including collision, out-of-lane, and freezing. We also evaluate the average velocity  (m/s) for efficiency evaluation. Full results, including additional metrics such as social compliance, are presented in the Appendix.





\vspace{-2mm}

\section{Experiments}
\vspace{-2mm}

We discuss the list of visual navigation models evaluated on \ourmethod in Sec.~\ref{sec:choice_models}. The results of \ourmethod including unit-test scenarios, pedestrian-reactive scenarios and long-horizon scenarios are presented in Sec.~\ref{sec:exp_unit_nav}, Sec.~\ref{sec:exp_ped_nav}, Sec.~\ref{sec:exp_long_nav}, respectively. Some qualitative results are shown in Fig.~\ref{fig:failure}.
To show the usefulness of our platform in generating synthetic data for model training, we conduct a preliminary finetuning experiment in Sec.~\ref{sec:exp_finetune} for the pedestrian-reactive scenarios.
\vspace{-2mm}
\subsection{Navigation Models}
\vspace{-2mm}
\label{sec:choice_models}
\begin{table}[t]
\centering
\footnotesize
\resizebox{\textwidth}{!}{%
\begin{tabular}{llllllll}
\toprule
\textbf{Method} 
& \textbf{Data (\# Hours)} 
& \textbf{Goal} 
& \textbf{Encoder} 
& \textbf{Decoder} 
& \textbf{Resolution} 
& \textbf{FPS}
& \textbf{Model Size} \\
\midrule

ViNT~\cite{shah2023vint}  
& General
& Image
& CNN
& Regression
& 85$\times$64
& 4
& 24~M \\

NoMaD~\cite{sridhar2024nomad} 
& General
& Image~/~None
& CNN
& Diffusion
& 96$\times$96
& 4
& 16~M \\

MBRA~\cite{hirose2025learning}  
& General
& Point
& CNN
& Regression
& 96$\times$96
& 5
& 63~M \\

CityWalker~\cite{liu2025citywalker} 
& General
& Point
& ViT
& Regression
& 630$\times$350
& 1
& 201~M \\

InternVLA-N1~\cite{wei2026ground} 
& General
& Language
& VLM
& Diffusion
& 384$\times$384
& 5
& 7~B \\

\midrule

MIMIC~\cite{he2026learning}  
& Sidewalk (50) 
& None
& ViT
& Reg.~+~Cls.
& 256$\times$256
& 5
& 79~M \\

S2E~\cite{he2025seeing}   
& Sidewalk (100)
& Point
& ViT
& Reg.~+~Cls.
& 256$\times$256
& 5
& 95~M \\

FlowPilot~\cite{he2026imitation}
& Sidewalk (300)
& Point~/~None
& ViT
& FM.~+~Cls.
& 512$\times$288
& 5
& 230~M \\

OpenPilot~\cite{openpilot}
& Sidewalk (1000)
& None
& ViT
& Regression
& 352$\times$128
& 20
& 8~M \\

\bottomrule
\end{tabular}%
}

\vspace{4pt}
\caption{\textbf{Detailed comparison of visual navigation models evaluated on \ourmethod.}}
\vspace{-8mm}
\label{tab:model_stats}
\end{table}

We evaluate 9 recent visual navigation models on our benchmark. We choose ViNT~\cite{shah2023vint}, NoMaD~\cite{sridhar2024nomad}, 
MBRA~\cite{hirose2025learning}, and CityWalker~\cite{liu2025citywalker} as representative models for visual navigation which are trained on diverse data sources. 
We further evaluate InternVLA-N1~\cite{wei2026ground} as VLA-based general navigation model with language inputs.
We also include  S2E~\cite{he2025seeing}, MIMIC~\cite{he2026learning}, FlowPilot~\cite{he2026imitation}, and OpenPilot~\cite{openpilot} as recent sidewalk navigation models trained specifically on data collected by human-teleoperated robot fleets navigating sidewalks across different cities. Detailed statistics of the models can be found in Tab~\ref{tab:model_stats}. Besides training data, the models also differ in architectures like vision encoders and action decoders as well as  input frequency and model size. This allows us to examine the roles that both data and architecture play in the benchmark performance. For example, OpenPilot~\cite{openpilot} has the smallest size of only 8M parameters, yet its observation frequency is the highest with 20 FPS.  Conversely, the VLA-based InternVLA-N1~\cite{wei2026ground} is the largest at 7B parameters and operates at a much slower rate of 5 FPS. More details of the models are presented in Appendix~\ref{supp:nav_scenario}.

\vspace{-2mm}
\subsection{Unit-test Scenarios}
\begin{figure}[t]
    \centering
    \includegraphics[width=1.0\textwidth]{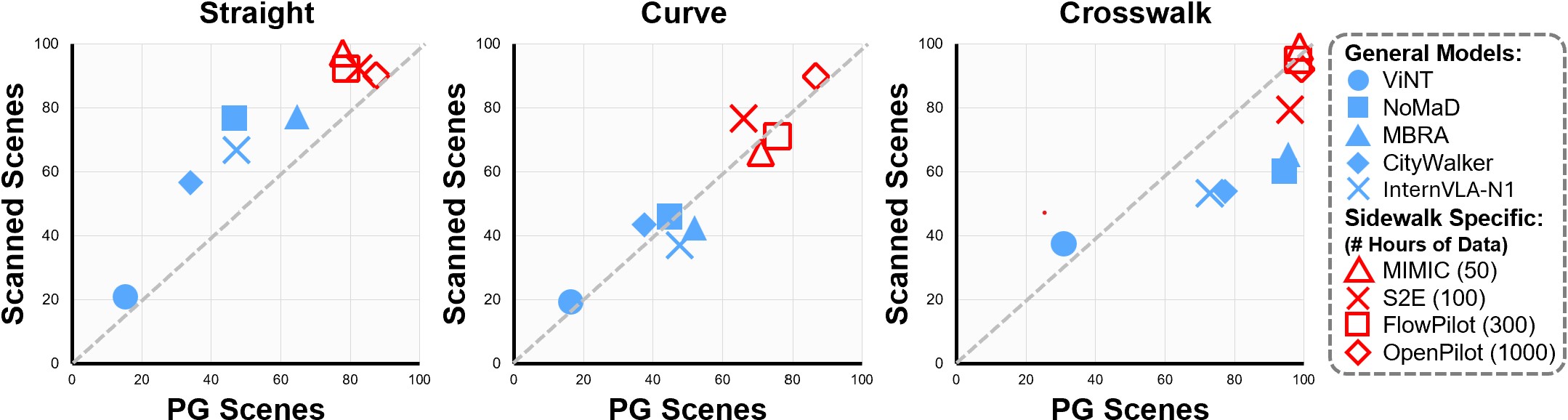}
    \vspace{-5mm}
    \caption{\textbf{Evaluation results of unit-test scenarios.} We show route completion rate (\%)   on different sidewalk structures in both procedurally generated (PG) and real-world scanned scenes. }
     \vspace{-2mm}
    \label{fig:unit_test}
\end{figure}

\vspace{-2mm}
We show results of the unit test scenarios on different topological structures of sidewalks in Fig.~\ref{fig:unit_test}. Overall, the general visual navigation foundation models perform much worse than the models trained on sidewalk-specific data. For example, ViNT~\cite{shah2023vint}, which is trained on diverse data sources, achieves 15.4~\% and 20.8~\% route completion rate on the straight blocks of procedurally generated scenes and real-world scanned scenes, while MIMIC~\cite{he2026learning} with only 50 hours of sidewalk training data achieves a better performance with 77.9~\% and 97.1~\% route completion rate in the same setting. This underlines the importance of sidewalk-specific data in model training. 

Moreover, the data-scaling effect applies to sidewalk navigation. For instance, FlowPilot~\cite{he2026imitation}, trained on 300 hours of data, outperforms MIMIC and achieves a route completion of 75.9~\% and 71.0~\% on the curve blocks. This is further surpassed by OpenPilot~\cite{openpilot}, which achieves a 86.5~\% and 89.8~\% average route completion rate using 1,000 hours of training data. To isolate the
effect of model architecture, we further conduct ablation experiments on the training data in Appendix~\ref{supp:data_ablation}.
%
%

Fig.~\ref{fig:unit_test} also demonstrates a strong correlation between performance in procedurally generated scenes and real-world scanned scenes. This implies that despite the real-to-sim visualization gap, procedurally generated environments remain highly useful for evaluating a model's real-world deployment performance.
More results  are presented in Appendix~\ref{supp:unit_test}.


\label{sec:exp_unit_nav}

\begin{figure}[t]
    \centering
    \includegraphics[width=1.0\textwidth]{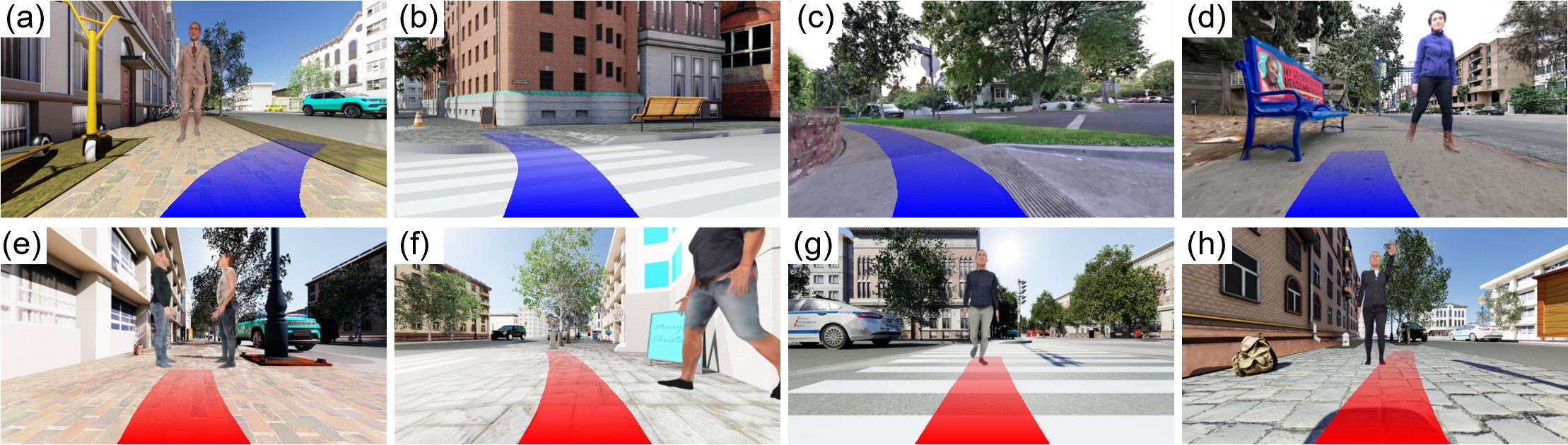}
    \vspace{-4mm}
    \caption{\textbf{Qualitative results}. (a)-(d): Success examples. (a) Avoiding a frontal approaching pedestrian. (b) Identifying the ramp. (c) Following the sidewalk curve. (d) Narrow passing. (e)-(h): Typical failure cases. (e) Ignoring social groups. (f) Failure to yield to a laterally crossing pedestrian. (g) Colliding with pedestrians on the crosswalk. (h) Inability to interpret pedestrian gestures.}
    \label{fig:failure}
     \vspace{-4mm}

\end{figure}
\vspace{-2mm}
\subsection{Pedestrian-reactive Scenarios}
\vspace{-2mm}
As shown in Tab.~\ref{tab:hand_crafted}, we evaluate the models' performance under different sidewalk pedestrian behaviors introduced in Sec.~\ref{sec:ped_simulation}.  From the comparison between the  `Static' scenario and the  `Obstructing' scenario, we can see that even a single static pedestrian could bring significant challenge to most models, with the average success rate dropping from 37.2~\% to 30.6~\%. From the comparison between the `Obstruction' scenario and the `Conversing' and `Queuing' scenarios, we observe that identifying social groups introduces additional challenges, and the average success rate drops further from 30.6~\% to 17.2~\% and 12.2~\%, respectively. 

Next, the ‘Frontal’, ‘Lateral’, and ‘Overtaking’ scenarios represent the three possible directions in which a pedestrian can approach the robot. The comparison results indicate that the frontal and lateral directions are the most challenging (averaging 17.3~\% and 19.4~\% success rates, respectively), requiring the robot to yield to a non-cooperative pedestrian promptly to avoid collisions. In contrast, the overtaking direction is significantly less challenging (averaging 43.2~\%), as models generally have sufficient time to yield to pedestrians approaching from behind.

The `Ped-Crossing' scenario ends up being the most challenging scenario with only a 3.8~\% success rate. This highlights the significant challenge of compounding factors: pedestrians approaching from multiple directions, adhering to the crosswalk lane, and locating the sidewalk ramp. The `Gesturing' scenario is almost equally challenging, with most methods failing to stop before the pedestrian. This underlines the lack of human gesture understanding with existing models, including the VLA-based model InternVLA-N1~\cite{wei2026ground}. Some common failure modes are visualized in Fig.~\ref{fig:failure}, and more results and analysis are presented in Appendix~\ref{supp:ped_reactive}.
   

\label{sec:exp_ped_nav}
\begin{table}[t]
\centering
\setlength{\tabcolsep}{2.0pt}
\definecolor{darkblue}{rgb}{0.75,0.85,0.95}
\definecolor{midblue}{rgb}{0.85,0.92,0.98}
\definecolor{lightblue}{rgb}{0.92,0.96,1.0}
\footnotesize
\resizebox{\textwidth}{!}{%
\begin{tabular}{@{}lccccccccc@{}}
\toprule
\textbf{Method} & \textbf{Static} & \textbf{Obstructing} & \textbf{Conversing} & \textbf{Queueing} & \textbf{Frontal} & \textbf{Lateral} & \textbf{Overtaking} & \textbf{Ped-Crossing} & \textbf{Gesturing} \\
\midrule
ViNT~\cite{shah2023vint} & $0.6_{\pm 0.9}$ & $0.2_{\pm 0.4}$ & $0.0_{\pm 0.0}$ & $0.2_{\pm 0.4}$ & $0.6_{\pm 0.9}$ & $0.6_{\pm 0.5}$ & $0.0_{\pm 0.0}$ & $0.6_{\pm 1.3}$ & $0.0_{\pm 0.0}$ \\
NoMaD~\cite{sridhar2024nomad} & $14.4_{\pm 3.5}$ & $13.4_{\pm 5.0}$ & $8.6_{\pm 1.1}$ & $4.4_{\pm 2.5}$ & $13.8_{\pm 2.5}$ & $3.4_{\pm 1.1}$ & $10.4_{\pm 2.6}$ & $4.6_{\pm 2.1}$ & $0.0_{\pm 0.0}$ \\
MBRA~\cite{hirose2025learning} & $32.4_{\pm 9.9}$ & $19.2_{\pm 2.0}$ & $9.4_{\pm 1.9}$ & $6.0_{\pm 2.8}$ & \cellcolor{lightblue}$19.0_{\pm 3.2}$ & $4.0_{\pm 1.2}$ & $15.8_{\pm 2.4}$ & \cellcolor{lightblue}$5.8_{\pm 1.9}$ & $0.0_{\pm 0.0}$ \\
CityWalker~\cite{liu2025citywalker} & $2.0_{\pm 1.2}$ & $2.6_{\pm 1.1}$ & $1.0_{\pm 0.7}$ & $1.4_{\pm 0.5}$ & $0.6_{\pm 0.9}$ & $0.8_{\pm 0.8}$ & $8.6_{\pm 3.5}$ & \cellcolor{midblue}$7.2_{\pm 2.0}$ & \cellcolor{darkblue}$11.6_{\pm 3.2}$ \\
InternVLA-DualVLN~\cite{wei2026ground} & $20.0_{\pm 3.7}$ & $17.8_{\pm 4.1}$ & $12.8_{\pm 3.5}$ & $6.0_{\pm 3.2}$ & $18.8_{\pm 3.3}$ & $3.2_{\pm 1.6}$ & $11.4_{\pm 2.1}$ & $1.2_{\pm 1.1}$ & $0.0_{\pm 0.0}$ \\
\midrule
MIMIC~\cite{he2026learning} & $60.2_{\pm 6.5}$ & $31.4_{\pm 2.3}$ & $5.2_{\pm 2.0}$ & $2.8_{\pm 1.5}$ & $2.6_{\pm 1.1}$ & \cellcolor{darkblue}$65.2_{\pm 6.1}$ & \cellcolor{midblue}$93.6_{\pm 2.3}$ & $0.4_{\pm 0.9}$ & $0.0_{\pm 0.0}$ \\
S2E~\cite{he2025seeing} & \cellcolor{midblue}$68.4_{\pm 2.7}$ & \cellcolor{midblue}$63.0_{\pm 5.2}$ & \cellcolor{midblue}$36.4_{\pm 4.4}$ & \cellcolor{lightblue}$22.2_{\pm 3.8}$ & \cellcolor{darkblue}$43.8_{\pm 9.4}$ & \cellcolor{midblue}$46.8_{\pm 3.8}$ & \cellcolor{lightblue}$80.0_{\pm 3.9}$ & $1.0_{\pm 1.0}$ & \cellcolor{lightblue}$0.2_{\pm 0.4}$ \\
FlowPilot~\cite{he2026imitation} & \cellcolor{lightblue}$66.8_{\pm 3.6}$ & \cellcolor{lightblue}$60.2_{\pm 4.1}$ & \cellcolor{lightblue}$33.6_{\pm 7.2}$ & \cellcolor{midblue}$32.4_{\pm 5.0}$ & \cellcolor{midblue}$40.6_{\pm 5.3}$ & $19.8_{\pm 3.1}$ & $74.0_{\pm 4.1}$ & \cellcolor{darkblue}$13.6_{\pm 4.8}$ & $0.0_{\pm 0.0}$ \\
OpenPilot~\cite{openpilot} & \cellcolor{darkblue}$81.8_{\pm 3.4}$ & \cellcolor{darkblue}$83.0_{\pm 4.4}$ & \cellcolor{darkblue}$48.2_{\pm 5.5}$ & \cellcolor{darkblue}$35.0_{\pm 3.5}$ & $16.4_{\pm 5.6}$ & \cellcolor{lightblue}$34.8_{\pm 5.9}$ & \cellcolor{darkblue}$95.2_{\pm 3.7}$ & $0.0_{\pm 0.0}$ & \cellcolor{midblue}$3.8_{\pm 2.2}$ \\
\midrule
\textbf{Avg} & $38.5_{\pm 1.2}$ & $32.3_{\pm 1.5}$ & $17.2_{\pm 1.1}$ & $12.3_{\pm 0.8}$ & $17.4_{\pm 2.0}$ & $19.8_{\pm 0.7}$ & $43.2_{\pm 0.9}$ & $3.8_{\pm 0.4}$ & $1.7_{\pm 0.3}$ \\
\bottomrule
\end{tabular}
}
\vspace{2mm}
\caption{\textbf{Model success rates (\%) of pedestrian-reactive scenarios}.  \textbf{Static} denotes a scenario without pedestrians as reference. The top three positive ranks in each column are highlighted from dark to light blue; ties share a shade. \textbf{Avg} is the mean across all nine models.  We also report the standard deviation across 5 runs with different random seeds. }
\label{tab:hand_crafted}
\vspace{-4mm}
\end{table}

  


\begin{figure*}[t]
    \centering
    \begin{minipage}[t]{0.65\textwidth}
        \vspace{0pt} 
        \centering
        \includegraphics[width=\linewidth]{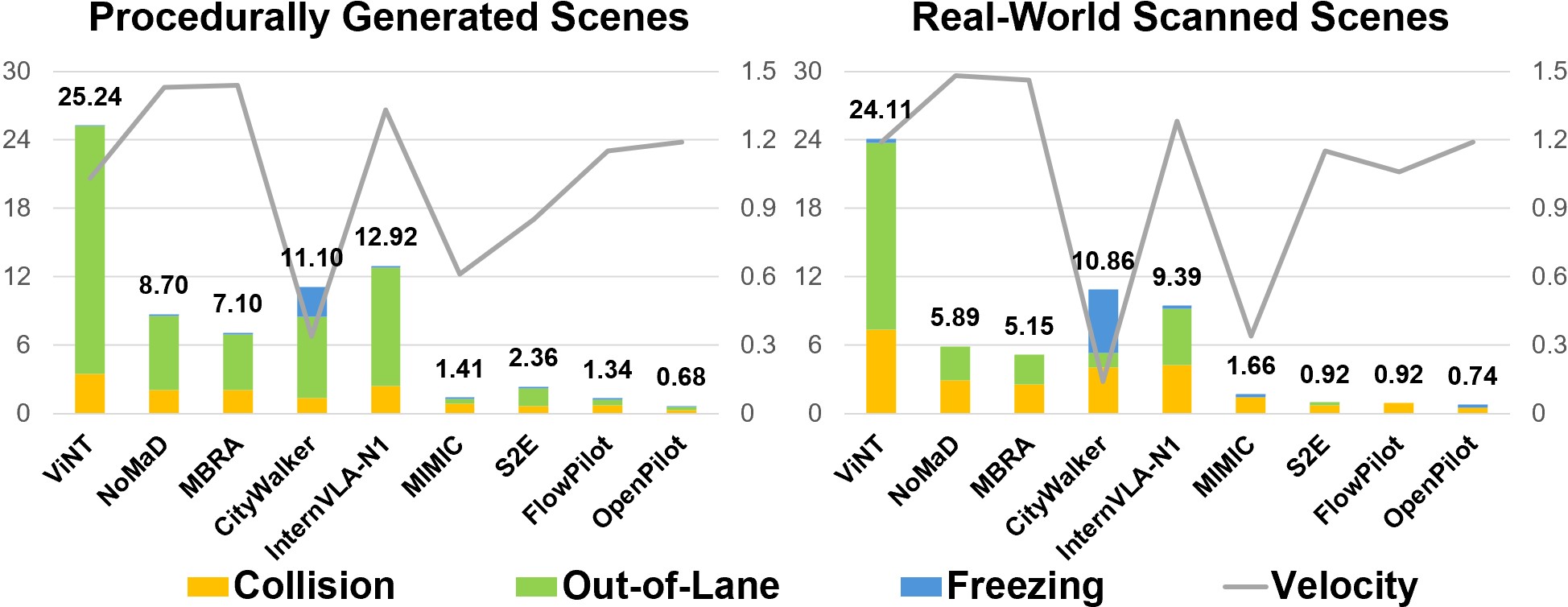}
        \vspace{-4mm} 
        \caption{\textbf{Evaluation results of long-horizon scenarios}.}
        \label{fig:long-horizon}
    \end{minipage}
    \hspace{0.03\textwidth} 
    \begin{minipage}[t]{0.30\textwidth}
        \vspace{0pt} 
        \centering
        {\footnotesize
        \setlength{\tabcolsep}{3.5pt}
        \resizebox{\linewidth}{!}{
        \begin{tabular}{l cc}
            \toprule
            \textbf{Scenario} & \textbf{Before} & \textbf{After} \\
            \midrule
            Ped-Crossing (Sim)  & 13.6 & \textbf{69.0} \\
            Ped-Crossing (Real) & 0.0 & \textbf{40.0} \\
            Gesturing (Sim)     & 0.0 & \textbf{49.0} \\
            Gesturing (Real)    & 0.0 & \textbf{50.0} \\
            \bottomrule
        \end{tabular}}}
        \captionof{table}{Success rates (\%) of FlowPilot~\cite{he2026imitation} before and after finetuning using our simulation platform for synthetic data generation.}
        \label{tab:finetune}
    \end{minipage}
\vspace{-6mm}
\end{figure*}
\vspace{-2mm}
\subsection{Long-horizon Scenarios}
\vspace{-2mm}
\begin{figure}[t]
    \centering
    \includegraphics[width=1.0\textwidth]{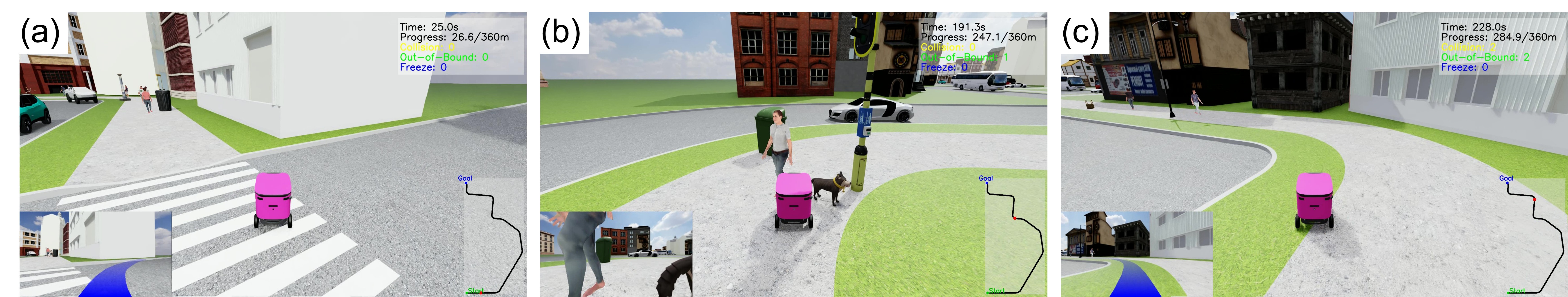}
    \caption{\textbf{Qualitative results of the long-horizon scenarios.} We visualize the failure cases of the two best performing models, FlowPilot~\cite{he2026imitation} and OpenPilot~\cite{openpilot}. (a) Failure to follow the crosswalk. (b) Collision when facing both a pedestrian and an obstacle in the front. (c) Out of the boundary while turning. We strongly encourage viewing these results  as videos on the project page.}
    \label{fig:quali_4}
 \vspace{-4mm}
\end{figure}
The long-horizon navigation performance of the evaluated models is presented in Fig.~\ref{fig:long-horizon}.  We can observe similar trends in procedurally-generated and real-world scanned scenes, and OpenPilot~\cite{openpilot}—trained on 1,000 hours of sidewalk navigation data—outperforms all the other models with the lowest failure rate of 0.68 per 100 meters and the  highest average speed among sidewalk-specific models of 1.19 m/s in procedurally-generated scenes. It is also worth noting that OpenPilot only uses a lightweight vision encoder paired with a simple regression-based action decoder~\cite{chen2022level}. These results underscore the importance of scaling sidewalk training data for robust long-horizon navigation, which outweighs other factors such as model architecture.  Nevertheless, our results imply that deploying OpenPilot on a food-delivery robot, which on average needs to drive 2,000 meters to reach the target~\cite{jennings2019study}, would still require 13.6 human takeovers—roughly 32.6 interventions per hour.  This demonstrates that a substantial gap remains for future research before fully autonomous operation can be achieved.  We visualize the failure cases of the two best performing models, FlowPilot~\cite{he2026imitation} and OpenPilot~\cite{openpilot} in the long-horizon scenarios in Fig.~\ref{fig:quali_4}. We can see that the failures in long-horizon scenarios combine those in the unit-test and pedestrian-reactive scenarios, and existing models still lack the robustness for long-horizon sidewalk navigation.

Figure~\ref{fig:long-horizon} further reveals distinct failure patterns across the two scene types. While most models
exhibit lower overall failure rates in real-world scanned scenes, their collision rates increase and
out-of-lane failures decrease relative to procedurally generated scenes. These differences suggest a
potential sim-to-real gap arising from variations in terrain geometry, camera observation, sidewalk
layouts, and obstacle configurations, and we leave further narrowing this gap to future work. More results are available in Appendix~\ref{supp:long_horizon}.
\label{sec:exp_long_nav}
\vspace{-2mm}
\subsection{Synthetic Training Data Generation}
\vspace{-2mm}
 \begin{figure}[t]
    \centering
    \includegraphics[width=1.0\textwidth]{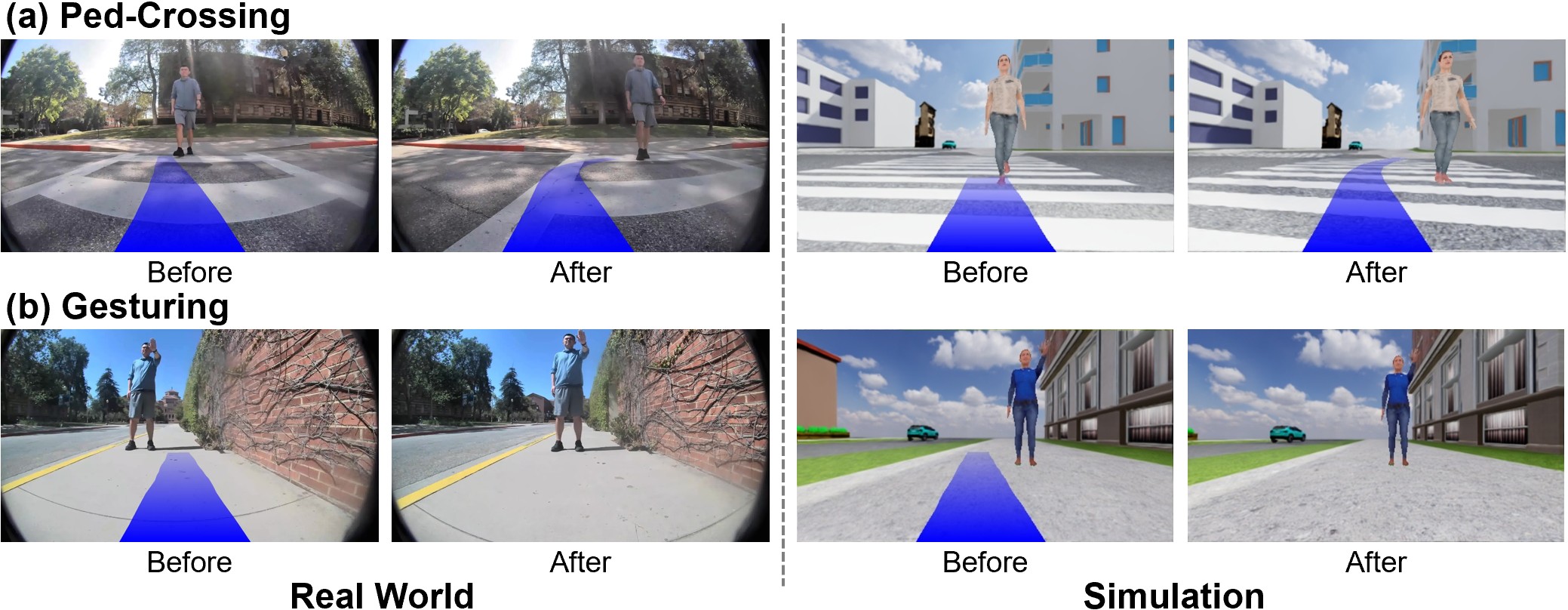}
    \vspace{-4mm}
    \caption{\textbf{Qualitative results of finetuning with synthetic data.} We visualize the model predictions of FlowPilot~\cite{he2026imitation} before and after finetuning using our simulation platform for synthetic
data generation. We strongly encourage viewing these results  as videos on the project page.}
     \vspace{-6mm}
    \label{fig:quali_5}
 
\end{figure}
\label{sec:exp_finetune}
As analyzed in the previous sections, scaling sidewalk training data is crucial for model performance. While collecting real-world pedestrian interaction data is expensive, our simulation platform can serve as a scalable synthetic data generator for model training. To this end, we conduct a preliminary experiment by fine-tuning on the two lowest-performing pedestrian-reactive scenarios: Ped-Crossing and Gesturing, which utilizes an expert planner to generate ground-truth robot actions for the FlowPilot~\cite{he2026imitation} model. The results shown in Tab.~\ref{tab:finetune} and Fig.~\ref{fig:quali_5} demonstrate that fine-tuning yields substantial performance gains in both simulated environments and the real world, showing the huge potential of scaling synthetic data to address current models' limitations in pedestrian behavior understanding and long-horizon robustness. More details  are presented in Appendix~\ref{supp:synthetic}.

\vspace{-2mm}

\section{Conclusion}
\vspace{-2mm}

We introduce~\ourmethod, a comprehensive benchmark for evaluating visual navigation on urban sidewalks. By simulating diverse and realistic sidewalk environments and further populating these environments with rich and reactive pedestrian behaviors and movements, \ourmethod allows standardized testing and comparison of different models. To reflect challenges in real-world sidewalk navigation, we further introduce the unit-test, pedestrian-reactive, and long-horizon testing scenarios and conduct a comprehensive  evaluation and analysis of representative visual navigation models.  Our findings show that current models suffer from pedestrian interaction and long-term robustness while scaling-up synthetic training data is a promising solution, and further research is needed before these models can be safely deployed in the real world.
\vspace{-4mm}

\section{Limitations} 
\vspace{-2mm}

First, our event-based pedestrian behaviors are governed by rule-based trajectories. While allowing standardized testing, these behaviors could lack realism compared to the more diverse and subtle robot-pedestrian interaction in the real world. Second, our procedurally-generated scenes trade off visual quality for efficiency and diversity, which could introduce rendering artifacts and lead to a larger real-to-sim gap. Third, the goal
of our experiments is to quantify the performance of visual navigation models and identify their failure modes.
Pinpointing the internals and reasoning about their failures
needs further study into each specific model. Lastly, the current real world experiments are at a relatively small scale aiming at demonstrating fine-tuning with synthetic data as a promising application, yet a broad sim-to-real conclusion needs additional validation.

\section*{Acknowledgment}
The project was supported by the ONR grants N000142512166 and N00014-23-1-2780, NSF grants CNS-2235012, TI-2346267, and CCF-2344955. We also acknowledge the hardware support from the NVIDIA Academic Grant Program.


\bibliography{ref}  
\newpage
\appendix
\suppressfloats[t]
\begin{center}{\Large\textbf{Appendix}}\end{center}

We present the implementation details of \ourmethod in Sec.~\ref{supp:nav_scenario}. To better support the claim that scaling sidewalk-specific training data is the most important factor for the model performance, we conduct additional ablation experiments in Sec.~\ref{supp:data_ablation}. More results and analysis of unit-test scenarios, pedestrian-reactive scenarios, and long-horizon scenarios are demonstrated in Sec.~\ref{supp:unit_test}, Sec.~\ref{supp:ped_reactive} and Sec.~\ref{supp:long_horizon}, respectively. We present additional details of synthetic training data generation in Sec.~\ref{supp:synthetic}.  More video results are available on the project page.
  

\begin{figure*}[t]
    \centering
    \begin{minipage}[t]{0.39\textwidth}
      \vspace{0pt} 
        \centering
        \includegraphics[width=\linewidth]{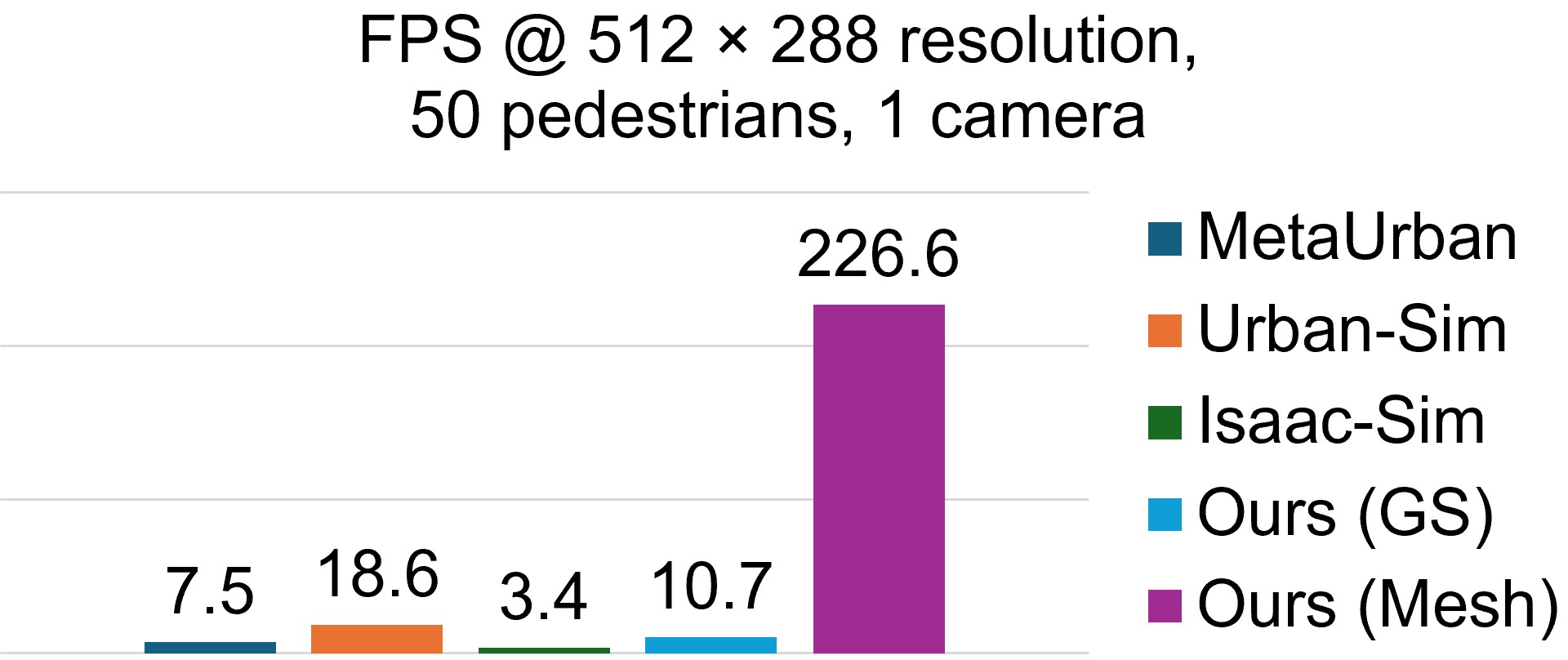}
         \caption{\textbf{Comparisons of the pedestrian rendering efficiency}. We report the amortized rendering FPS of pedestrians including motion generation, rigging, skinning, and rendering.}
        \label{fig:efficiency}
    \end{minipage}
    \hspace{0.00\textwidth} 
    \begin{minipage}[t]{0.59\textwidth}
        \vspace{0pt} 
        \centering
        \includegraphics[width=\linewidth]{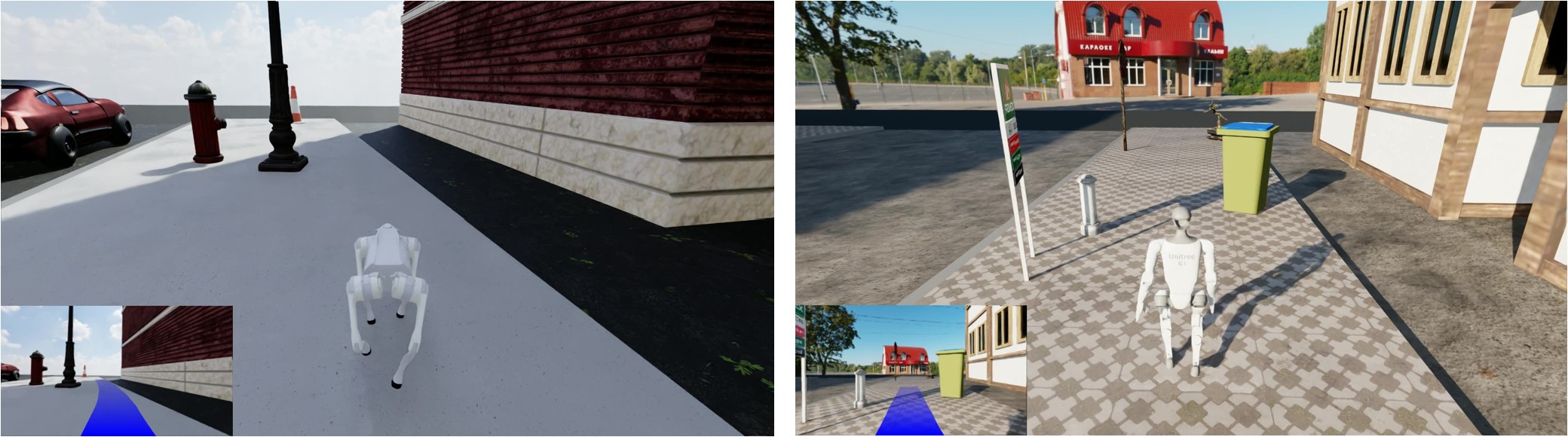}
         \caption{\textbf{Qualitative results of other robot embodiments}. We show examples of the Unitree Go2 robot dog (left) and the Unitree G1 humanoid (right) evaluated with the FlowPilot~\cite{he2026imitation} model.  Video results are available on the project page.}
        \label{fig:embodiment}
    \end{minipage}
\end{figure*}
\vspace{-2mm}
\section{Implementation Details}
\vspace{-2mm}

\label{supp:nav_scenario}

\paragraph{Pedestrian Simulation}
\label{supp:ped_simulation}
 \begin{table}[t]
\centering

\footnotesize
\resizebox{\textwidth}{!}{%
\begin{tabular}{@{}lll@{}}
\toprule
\textbf{Scenario} & \textbf{Description} & \textbf{Triggering Condition} \\
\midrule
\textbf{Obstructing}  & A stationary pedestrian stands at the center of the sidewalk looking around. &  Within 10m ahead of the robot \\
\textbf{Conversing} & Multiple pedestrians stand facing each other at the center of the path, forming a social group. &  Within 10m ahead of the robot \\
\textbf{Queueing}     & Pedestrians form a structured line starting from one side of the sidewalk with tight spacing. &  Within 10m ahead of the robot \\
\textbf{Frontal Approaching}      & A pedestrian walks directly toward the current position of the robot from the opposite direction. & Within 20m ahead of the robot \\
\textbf{Lateral Approaching} & A pedestrian crosses the robot's projected path laterally, requiring the robot to yield. & Within 4m ahead of the robot \\
\textbf{Overtaking}   & A pedestrian approaches and passes the robot from behind with a faster speed. & Within 1m behind the robot \\
\textbf{Ped-Crossing}     & One or more pedestrians crossing from front and behind across a designated crosswalk. & Robot enters the crosswalk zone \\

\textbf{Gesturing}      & A pedestrian continuously faces the robot while executing a distinct stopping gesture. & Within 10m ahead of the robot \\

\bottomrule
\end{tabular}%
}
\vspace{2mm}

\caption{\textbf{Descriptions of event-based pedestrian behaviors.}}
  \vspace{-5mm}
\label{tab:scenario_framework}
\end{table}
 The event-based behaviors are primarily used in pedestrian-reactive scenarios, with detailed descriptions listed in Tab.~\ref{tab:scenario_framework}.
 Note that both the trajectory and the speed of the pedestrian are specified by hand-crafted rules in  event-based behaviors  to ensure the robot has to react correspondingly to the pedestrian.
 For example, in the scenario "Lateral Approaching", we control the trajectory of the pedestrian so it is perpendicular to the current robot heading and adjust the speed so the robot would collide with the pedestrian if it keeps the current velocity.
 For pedestrian behaviors in the long-horizon scenarios like walking towards a target point,  we use ORCA~\cite{van2011reciprocal} to simulate reciprocal response to the robot’s maneuver with the pedestrian speed randomly sampled  from common walking speed (1.1~m/s~--~1.65~m/s). The pedestrian-reactive
scenarios are complementary to the long-horizon scenarios
with standardized testing of the model’s reaction to different
pedestrian behaviors. Though less common in the real world,
these event-triggered behaviors help identify the models’ failure modes to enhance safe interaction with pedestrians.

 For the local pedestrian movements, we use an efficient motion generation model Motion-LCM~\cite{dai2024motionlcm} to generate common pedestrian movements such as walking and standing with high diversity on-the-fly during simulation.
 For less common movements such as a stopping gesture, we use existing human motion datasets such as Motion-X~\cite{lin2023motion} to retrieve the corresponding motion.
 As mentioned in the main paper, our pedestrian animation is highly efficient, and a comparison of the rendering efficiency is shown in Fig.~\ref{fig:efficiency}, which shows the amortized rendering FPS including motion generation, rigging, skinning, and final rendering. 
 As our pedestrian animation is controlled by the SMPL model~\cite{loper2023smpl}, we developed two different rendering approaches.
 For the procedurally-generated scenes, we use mesh-based rendering that applies SMPL-compatible human textures~\cite{casas2023smplitex} to the SMPL body mesh. It can achieve a remarkable 226.6 FPS in a single environment with 50 pedestrians, which is much higher than prior works, including the native human animation pipeline~\cite{nvidia_isaac_sim_actor_control} in Isaac Sim with only a 3.4 FPS. This drastic improvement facilitates highly efficient simulation and evaluation of large, crowded urban scenes.
 For the real-world scanned scenes, we use SMPL 3DGS avatars~\cite{kocabas2024hugs, jiang2022neuman} with better fidelity and are more compatible with the 3DGS background scene. Our GS-based pedestrian rendering can still achieve a 10.7 FPS in the same setting and is comparable to other simulators like MetaUrban~\cite{wu2025metaurban} with a 7.5 FPS.
Since we use an independent pedestrian simulation pipeline, the pedestrian collision detections are also handled separately in our simulation platform, where we use a cylinder of radius 0.3m to represent the collision mesh of pedestrians to improve efficiency. This is reasonable for visual navigation that does not involve physical interactions with humans.

\paragraph{Testing Scenario Configurations}
\label{supp:nav_tasks}
We use a four-wheeled delivery robot~\cite{cocorobotics2026} as the main robot platform for our benchmark due to its simple dynamics and strong practicality in the real world.
The robot has a maximum speed of 1.5 m/s and a maximum angular velocity of 0.65 rad/s.
Our simulation platform also supports the evaluation on other robot embodiments such as the robot dog or the humanoid (shown in Fig.~\ref{fig:embodiment}). Following existing works~\cite{shah2023vint, he2025seeing, liu2025citywalker}, we use the same PD controller to convert the robot waypoints predicted by the model into uniform velocity commands including the linear and the angular velocity. We handpick textures and assets in UrbanVerse-100K~\cite{liu2026urbanverse} with high realism to reduce the sim-to-real gap in procedurally-generated scenes. To ensure reproducible results, we used synchronized testing, i.e. the simulator would wait for a robot velocity command at each step and ignore the inference latency. It is worth noting that the models do have large discrepancies in inference speed and could lead to different results in real-world deployment as shown in Tab.~\ref{tab:model_stats}. The benchmark results could also largely depend on the scene textures and the robot low-level controllers which we leave exploring their effects to the future work. All benchmarking experiments are conducted in 10 parallel environments on a NVIDIA L40S GPU in Isaac-Sim 6.0~\cite{NVIDIA_Isaac_Sim}. The physics simulation step is set to 0.005s to ensure high physics realism.
For the `Gesturing' scenario, the success criteria is the robot stops within 3m in front of the pedestrian instead of reaching the goal.
For the `Conversing' and the `Queueing' scenarios, we only count success when the robot avoids the whole pedestrian group instead of passing between the pedestrians.
%


\paragraph{Models}
\label{supp:model_stats}
%
During testing, we adjust the inference frequency and the image resolution of simulation to match the specifications of each model. 
Our evaluation protocol ensures a fair comparison between goal-conditioned models and goal-free models by
removing the privileged goal information. For image-goal
models, we evaluate ViNT and NoMaD in goal-free mode as
they already support goal-free exploration (ViNT-R explores
under noisy goals and NoMaD uses 50\% goal masking during
training). For point-goal models, we continuously update the
goal to be a random point in front of the ego robot with the longitudinal distance sampled from $(2, 10)$ meters and the lateral offset sampled from $(-2, 2)$ meters, so it
will not introduce bias toward shortcutting the provided goals. 
Such a setting is also practical in real-world deployment, as
goal images are often not available and the GPS for point goals
is noisy. 
For goal-free models, we use visual observations as the only input. 
For the VLA models, we use a goal-agnostic navigation instruction, ``Follow the sidewalk'' to avoid leaking task-specific information.  
This protocol allows each model to operate under its intended interface while reducing unfair advantages from explicit future information. Our released codebase also supports  evaluating goal-conditioned
models with route-consistent goal points or goal images. 
The OpenPilot model is re-implemented based on the public checkpoint~\cite{openpilot} and re-trained on 1000 hours of sidewalk data. It takes the recent two image frames and a historical context vector, and uses a recurrent neural network to output predicted future waypoints~\cite{chen2022level}.

\begin{table}[t]
\centering
 \resizebox{\columnwidth}{!}{%
\renewcommand{\arraystretch}{0.9 }  
\begin{tabular}{@{}lll|cccc|c@{}}
\toprule
\textbf{Row} & \textbf{Model} & \textbf{Data (\# Hours)} & \textbf{Straight} $\uparrow$ & \textbf{Curve} $\uparrow$ & \textbf{Crosswalk} $\uparrow$ & \textbf{Average} $\uparrow$ & \textbf{Long-horizon} $\downarrow$ \\
\midrule
1 & ViNT~\cite{shah2023vint} & General & 15.4 & 16.5 & 30.9 & 20.9 & 25.24 \\
2 & ViNT~\cite{shah2023vint} & Sidewalk (300) & 72.6 & 85.8 & 99.4 & 85.9 & 1.24 \\
\midrule
3 & OpenPilot~\cite{openpilot} & Sidewalk (10) & 68.2 & 53.2 & 99.8 & 73.7 & 7.57 \\
4 & OpenPilot~\cite{openpilot} & Sidewalk (100) & 78.0 & 72.3 & 95.4 & 81.9 & 2.59 \\
5 & OpenPilot~\cite{openpilot} & Sidewalk (300) & 81.1 & 78.0 & 95.2 & 84.8 & 1.80 \\
6 & OpenPilot~\cite{openpilot} & Sidewalk (1000) & 87.6 & 86.5 & 99.3 & 91.1 & 0.68 \\
\bottomrule
\end{tabular}
}
 \vspace{2mm}

\caption{\textbf{Ablation on scaling training data.} We report the route completion rate (\%) on three types of unit-test scenarios and their average, and the average failure/100m of long-horizon scenarios in the procedurally generated scenes.  
}

\label{tab:training_data}
\end{table}

\section{Ablation on Sidewalk Training Data} 
\vspace{-2mm}
\label{supp:data_ablation}
To isolate the effect of model architecture, we retrain OpenPilot~\cite{openpilot} with 10, 100 and 300 hours of sidewalk dataset and ViNT~\cite{shah2023vint} with the same 300 hours of data. The results are shown in Tab.~\ref{tab:training_data}. Comparing rows 1-2 shows training with sidewalk-specific data outperforms general data significantly. Comparing rows 3-6 shows that scaling sidewalk-specific data results in consistent performance gains. Comparing rows 2 vs. 5 reveals that the model architecture only brings marginal performance differences given the same training data. These results thus confirm that data scale and domain dominate architecture for sidewalk navigation. 

\section{More Results of Unit-test Scenarios}
 \vspace{-2mm}

\begin{figure}[t]
    \centering
      \vspace{-5mm}
    \includegraphics[width=0.7\textwidth]{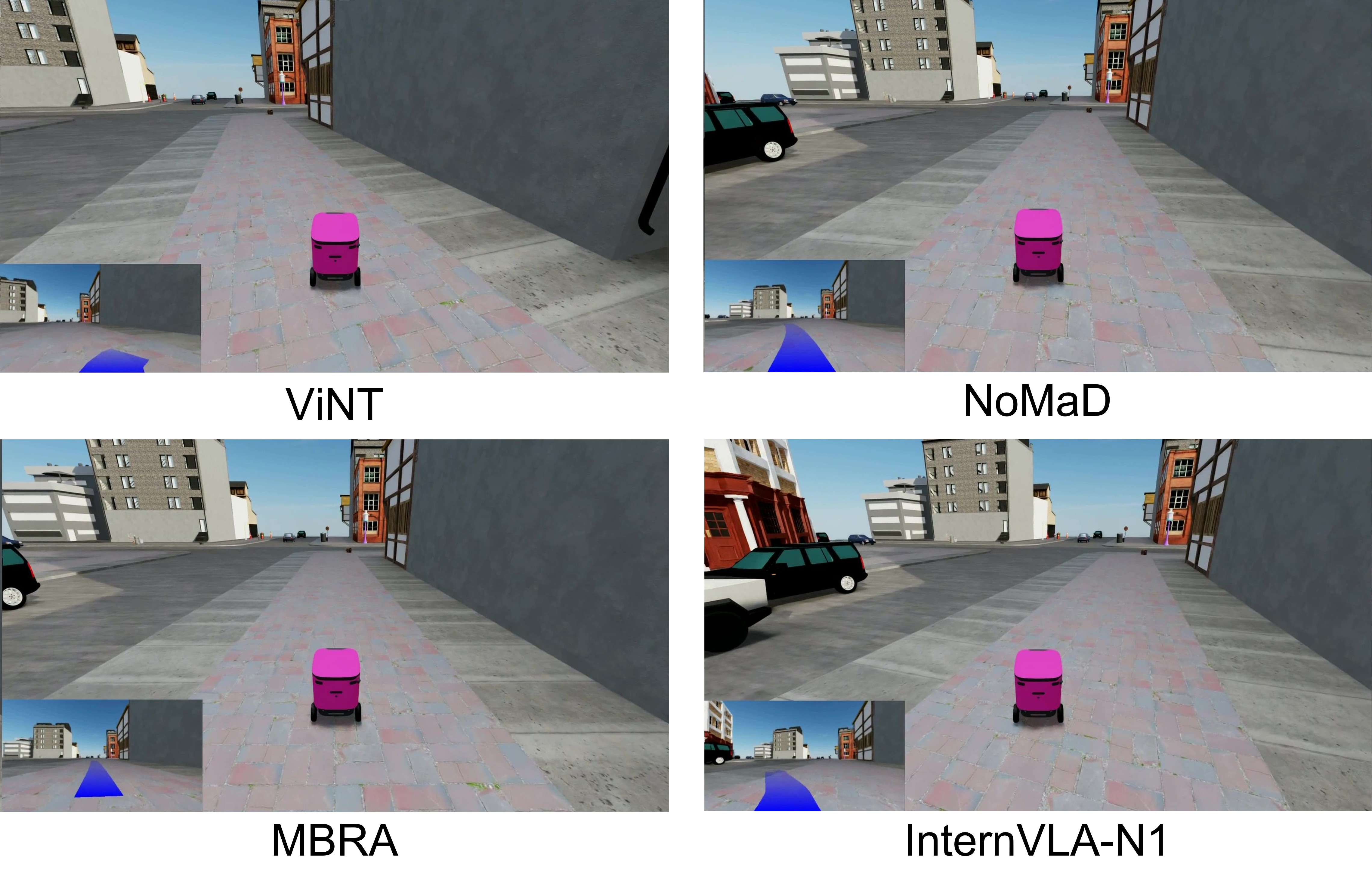}
    \vspace{-3mm}
    \caption{\textbf{Qualitative comparisons of the unit-test scenarios.} ViNT~\cite{shah2023vint}, NoMaD~\cite{sridhar2024nomad}, and InternVLA-N1~\cite{wei2026ground} are not able to follow the straight sidewalk lane. Only MBRA~\cite{hirose2025learning} has some basic lane-following ability. We strongly encourage viewing these results (and
other models) as videos on the project page.}
    \label{fig:quali_1}
 
\end{figure}

\label{supp:unit_test}
\begin{table}[t]
\centering
\footnotesize
\setlength{\tabcolsep}{1.5pt}
\resizebox{\textwidth}{!}{%
\begin{tabular}{@{}lccccccccc@{}}
\toprule
& \multicolumn{9}{c}{\textbf{Procedurally Generated Scenes}} \\
\cmidrule(lr){2-10}
 & \multicolumn{3}{c}{\textbf{Straight}} & \multicolumn{3}{c}{\textbf{Curve}} & \multicolumn{3}{c}{\textbf{Crosswalk}} \\
\cmidrule(lr){2-4} \cmidrule(lr){5-7} \cmidrule(lr){8-10}
\textbf{Method} & \textbf{SR\%}~$\uparrow$ & \textbf{SPL\%}~$\uparrow$ & \textbf{RC\%}~$\uparrow$ & \textbf{SR\%}~$\uparrow$ & \textbf{SPL\%}~$\uparrow$ & \textbf{RC\%}~$\uparrow$ & \textbf{SR\%}~$\uparrow$ & \textbf{SPL\%}~$\uparrow$ & \textbf{RC\%}~$\uparrow$ \\
\midrule
ViNT~\cite{shah2023vint} & $0.6_{\pm 0.9}$ & $0.6_{\pm 0.8}$ & $15.4_{\pm 1.3}$ & $0.6_{\pm 1.3}$ & $0.6_{\pm 1.3}$ & $16.5_{\pm 3.7}$ & $1.4_{\pm 1.1}$ & $1.4_{\pm 1.1}$ & $30.9_{\pm 3.1}$ \\
NoMaD~\cite{sridhar2024nomad} & $14.4_{\pm 3.5}$ & $13.9_{\pm 3.4}$ & $46.6_{\pm 2.5}$ & $11.2_{\pm 4.3}$ & $11.2_{\pm 4.3}$ & $44.8_{\pm 3.2}$ & $26.6_{\pm 7.9}$ & $25.6_{\pm 7.6}$ & $94.4_{\pm 0.8}$ \\
MBRA~\cite{hirose2025learning} & $32.4_{\pm 9.9}$ & $31.5_{\pm 9.6}$ & $64.7_{\pm 4.7}$ & $0.2_{\pm 0.4}$ & $0.2_{\pm 0.4}$ & $51.9_{\pm 0.8}$ & $35.2_{\pm 3.9}$ & $34.1_{\pm 3.8}$ & $95.5_{\pm 1.2}$ \\
CityWalker~\cite{liu2025citywalker} & $2.0_{\pm 1.2}$ & $1.9_{\pm 1.2}$ & $34.1_{\pm 2.3}$ & $1.6_{\pm 1.8}$ & $1.6_{\pm 1.8}$ & $37.5_{\pm 2.8}$ & $12.6_{\pm 4.3}$ & $12.1_{\pm 4.1}$ & $77.4_{\pm 4.3}$ \\
InternVLA-DualVLN~\cite{wei2026ground} & $20.0_{\pm 3.7}$ & $19.4_{\pm 3.6}$ & $47.2_{\pm 3.6}$ & $3.0_{\pm 1.6}$ & $3.0_{\pm 1.6}$ & $47.7_{\pm 1.3}$ & $3.2_{\pm 1.3}$ & $3.1_{\pm 1.3}$ & $72.8_{\pm 1.8}$ \\
\midrule
MIMIC~\cite{he2026learning} & $60.2_{\pm 6.5}$ & $60.2_{\pm 6.5}$ & $77.9_{\pm 2.5}$ & $43.6_{\pm 4.7}$ & $43.6_{\pm 4.7}$ & $71.0_{\pm 2.4}$ & $76.8_{\pm 5.0}$ & $76.8_{\pm 5.0}$ & $98.8_{\pm 1.2}$ \\
S2E~\cite{he2025seeing} & $68.4_{\pm 2.7}$ & $67.2_{\pm 2.6}$ & $82.4_{\pm 1.9}$ & $33.4_{\pm 5.0}$ & $33.0_{\pm 5.1}$ & $65.9_{\pm 2.1}$ & $63.2_{\pm 6.8}$ & $61.6_{\pm 6.6}$ & $95.9_{\pm 0.9}$ \\
FlowPilot~\cite{he2026imitation} & $66.8_{\pm 3.6}$ & $65.7_{\pm 3.6}$ & $79.0_{\pm 3.0}$ & $55.4_{\pm 4.3}$ & $55.3_{\pm 4.3}$ & $75.9_{\pm 3.5}$ & $71.8_{\pm 5.9}$ & $69.7_{\pm 5.7}$ & $98.3_{\pm 1.0}$ \\
OpenPilot~\cite{openpilot} & $\mathbf{81.8}_{\pm 3.4}$ & $\mathbf{81.3}_{\pm 3.4}$ & $\mathbf{87.6}_{\pm 3.3}$ & $\mathbf{71.4}_{\pm 6.8}$ & $\mathbf{71.3}_{\pm 6.8}$ & $\mathbf{86.5}_{\pm 4.3}$ & $\mathbf{95.2}_{\pm 1.8}$ & $\mathbf{94.3}_{\pm 1.8}$ & $\mathbf{99.3}_{\pm 1.0}$ \\
\bottomrule
\end{tabular}%
}
\vspace{2mm}
\caption{\textbf{Full evaluation results of unit-test scenarios on procedurally-generated scenes.} We evaluate success rate
(SR\%), success weighted by path length (SPL\%), and route completion rate (RC\%) on different
sidewalk structures. We also report the  standard deviation across 5 runs with different random seeds. }
\label{tab:terrain_only_pg}
\end{table}

\begin{table}[t]
  \centering
  \footnotesize
  \setlength{\tabcolsep}{1.5pt}
  \resizebox{\textwidth}{!}{%
  \begin{tabular}{@{}lccccccccc@{}}
  \toprule
  & \multicolumn{9}{c}{\textbf{Real-World Scanned Scenes}} \\
  \cmidrule(lr){2-10}
   & \multicolumn{3}{c}{\textbf{Straight}} & \multicolumn{3}{c}{\textbf{Curve}} & \multicolumn{3}{c}{\textbf{Crosswalk}} \\
  \cmidrule(lr){2-4} \cmidrule(lr){5-7} \cmidrule(lr){8-10}
  \textbf{Method} & \textbf{SR\%}~$\uparrow$ & \textbf{SPL\%}~$\uparrow$ & \textbf{RC\%}~$\uparrow$ & \textbf{SR\%}~$\uparrow$ &
  \textbf{SPL\%}~$\uparrow$ & \textbf{RC\%}~$\uparrow$ & \textbf{SR\%}~$\uparrow$ & \textbf{SPL\%}~$\uparrow$ &
  \textbf{RC\%}~$\uparrow$ \\
  \midrule
  ViNT~\cite{shah2023vint} & $0.0_{\pm 0.0}$ & $0.0_{\pm 0.0}$ & $20.8_{\pm 1.9}$ & $0.0_{\pm 0.0}$ & $0.0_{\pm 0.0}$ & $19.1_{\pm 3.1}$ & $0.0_{\pm 0.0}$ & $0.0_{\pm 0.0}$ &
  $37.4_{\pm 2.1}$ \\
  NoMaD~\cite{sridhar2024nomad} & $52.0_{\pm 8.4}$ & $50.7_{\pm 8.3}$ & $76.6_{\pm 4.4}$ & $4.0_{\pm 5.5}$ & $3.9_{\pm 5.3}$ & $46.0_{\pm 3.7}$ & $18.0_{\pm 4.5}$ & $17.7_{\pm 4.4}$ &
  $59.9_{\pm 2.1}$ \\
  MBRA~\cite{hirose2025learning} & $52.0_{\pm 8.4}$ & $51.3_{\pm 8.5}$ & $77.2_{\pm 3.1}$ & $0.0_{\pm 0.0}$ & $0.0_{\pm 0.0}$ & $42.5_{\pm 3.6}$ & $16.0_{\pm 13.4}$ & $15.8_{\pm 13.3}$
  & $65.1_{\pm 4.2}$ \\
  CityWalker~\cite{liu2025citywalker} & $14.0_{\pm 5.5}$ & $13.9_{\pm 5.4}$ & $56.5_{\pm 5.5}$ & $0.0_{\pm 0.0}$ & $0.0_{\pm 0.0}$ & $43.3_{\pm 5.3}$ & $0.0_{\pm 0.0}$ & $0.0_{\pm
  0.0}$ & $53.7_{\pm 9.1}$ \\
  InternVLA-N1~\cite{wei2026ground} & $36.0_{\pm 11.4}$ & $35.2_{\pm 10.7}$ & $66.7_{\pm 5.1}$ & $4.0_{\pm 5.5}$ & $4.0_{\pm 5.4}$ & $37.2_{\pm 3.0}$ & $14.0_{\pm 5.5}$ & $13.2_{\pm
  5.4}$ & $53.1_{\pm 1.4}$ \\
  \midrule
  MIMIC~\cite{he2026learning} & $84.0_{\pm 5.5}$ & $83.3_{\pm 5.5}$ & $\mathbf{97.1}_{\pm 0.4}$ & $30.0_{\pm 7.1}$ & $29.4_{\pm 7.0}$ & $66.0_{\pm 2.7}$ & $\mathbf{94.0}_{\pm 5.5}$ &
  $\mathbf{92.8}_{\pm 5.2}$ & $\mathbf{99.2}_{\pm 0.7}$ \\
  S2E~\cite{he2025seeing} & $\mathbf{90.0}_{\pm 0.0}$ & $88.8_{\pm 0.1}$ & $92.3_{\pm 0.0}$ & $54.0_{\pm 5.5}$ & $53.9_{\pm 5.5}$ & $76.5_{\pm 3.7}$ & $44.0_{\pm 8.9}$ & $43.3_{\pm
  8.8}$ & $79.3_{\pm 2.5}$ \\
  FlowPilot~\cite{he2026imitation} & $\mathbf{90.0}_{\pm 0.0}$ & $89.1_{\pm 0.1}$ & $92.2_{\pm 0.1}$ & $46.0_{\pm 8.9}$ & $45.3_{\pm 8.7}$ & $71.0_{\pm 4.2}$ & $90.0_{\pm 0.0}$ &
  $89.2_{\pm 0.0}$ & $94.9_{\pm 1.0}$ \\
  OpenPilot~\cite{openpilot} & $\mathbf{90.0}_{\pm 0.0}$ & $\mathbf{89.3}_{\pm 0.0}$ & $90.1_{\pm 0.0}$ & $\mathbf{82.0}_{\pm 4.5}$ & $\mathbf{81.8}_{\pm 4.5}$ & $\mathbf{89.8}_{\pm
  2.3}$ & $92.0_{\pm 4.5}$ & $91.4_{\pm 4.6}$ & $92.1_{\pm 4.4}$ \\
  \bottomrule
  \end{tabular}%
  }
\vspace{2mm}
\caption{\textbf{Full evaluation results of unit-test scenarios on real-world scanned scenes.} We evaluate success rate
(SR\%), success weighted by path length (SPL\%), and route completion rate (RC\%) on different
sidewalk structures. We also report the standard deviation across 5 runs with different random seeds. }
\label{tab:terrain_only_scanned}
\end{table}

Full evaluation results of the unit-test scenarios are shown in Tab.~\ref{tab:terrain_only_pg} and Tab.~\ref{tab:terrain_only_scanned} for the procedurally generated scenes and real-world scanned scenes respectively. We report success rate (SR) and success weighted by path length (SPL)~\cite{anderson2018evaluation} as additional metrics commonly used in visual navigation besides route completion rate (RC).  
We can see that the success rates for ViNT~\cite{shah2023vint}
are almost 0 under all settings, meaning it lacks basic lane following and obstacle avoidance abilities on urban sidewalks despite being a general foundation model for visual navigation.
In contrast, MBRA~\cite{hirose2025learning} achieves significantly better performance, with the highest SR and SPL of 32.4~\% and 31.5~\% among all the general navigation models on the straight blocks in procedurally-generated scenes. It is worth mentioning that the main difference between MBRA and the previous two models is that MBRA uses a specific data filtering procedure that ensures high label quality, which further highlights the importance of training data for sidewalk navigation.

\begin{wrapfigure}{r}{0.35\textwidth}
 \vspace{-4mm}

\centering
\includegraphics[width=\linewidth]{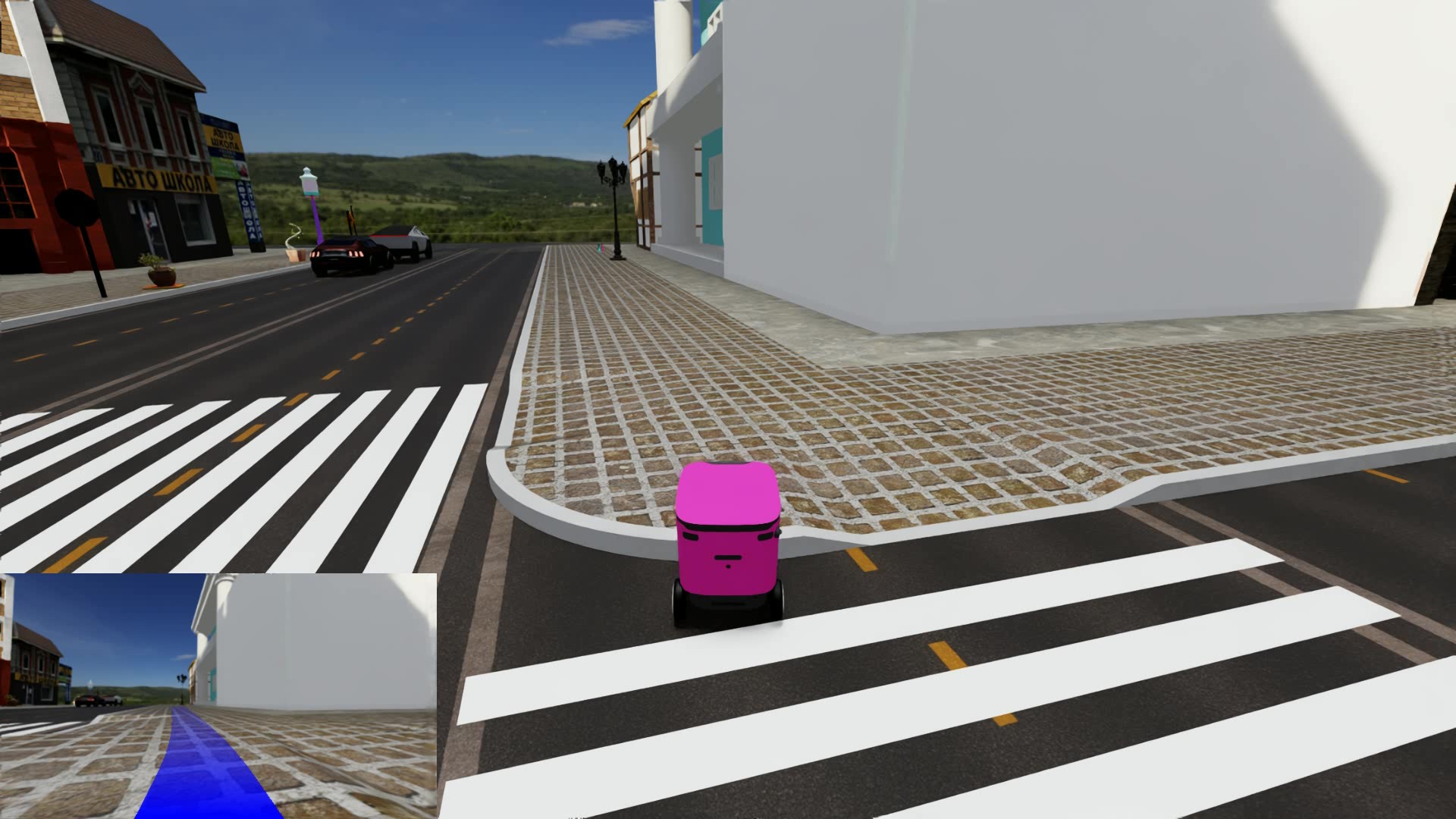}
 
\caption{Failure to identify the ramp at the end of the crosswalk.}
    \label{fig:quali_2}

\vspace{-5mm}
\label{fig:Attn}
 
\end{wrapfigure}

Meanwhile, despite the heavy model architecture and slower inference speed, InternVLA-N1~\cite{wei2026ground} still underperforms compared to MBRA and the sidewalk-specific models. This shows that while VLM backbones offer promising high-level reasoning capabilities, current VLA-based models fall short in unit-testing scenarios that demand more on timely and precise control which could be learned from sidewalk-specific training data. Some qualitative comparisons are shown in Fig.~\ref{fig:quali_1}.

We also   observe a minimal gap between SR and SPL, implying all the models achieve a relatively high path efficiency.
This is  expected  as sidewalk navigation requires the model to follow the lane without taking much detour except for avoiding obstacles, and thus the completion time weights more than the path length in the overall efficiency evaluation.
Among the three structures, the curve block is the most challenging, which requires the robot to follow a curved path and avoid static obstacles. For instance, the best model OpenPilot~\cite{openpilot} results in a 90.0~\%, 82.0~\% and 92.0~\% SR on the straight, curve, and crosswalk blocks respectively in real-world scanned scenes. While the crosswalk has the highest success rate because it has no obstacles, we observe identifying the ramp at the end of the crosswalk remains challenging for many models (illustrated in Fig.~\ref{fig:quali_2}), resulting in a high route completion rate   but a lower success rate. For example, S2E~\cite{he2025seeing} achieves a 95.9~\% route completion rate on crosswalks in procedurally generated scenes, but its success rate drops to only 63.2~\% in the same setting. 
\section{More Results of  Pedestrian-reactive Scenarios}
\label{supp:ped_reactive}
\begin{table}[t]
\centering
\setlength{\tabcolsep}{2.0pt}
\footnotesize
\resizebox{\textwidth}{!}{%
\begin{tabular}{@{}llcccccccc@{}}
\toprule
\textbf{Metric} & \textbf{Method} & \textbf{Obstructing} & \textbf{Conversing} & \textbf{Queueing} & \textbf{Frontal} & \textbf{Lateral} & \textbf{Overtaking} & \textbf{Ped-Crossing} & \textbf{Gesturing} \\
\midrule
\multirow{10}{*}{\shortstack[l]{\textbf{Pedestrian}\\\textbf{Collision}\\\textbf{Rate (\%)} $\downarrow$}} & ViNT~\cite{shah2023vint} & $0.8$ & $1.0$ & $2.2$ & $0.2$ & $1.2$ & $0.0$ & $3.6$ & $0.6$ \\
 & NoMaD~\cite{sridhar2024nomad} & $11.2$ & $16.6$ & $23.4$ & $8.2$ & $28.2$ & $14.0$ & $42.6$ & $9.4$ \\
 & MBRA~\cite{hirose2025learning} & $26.0$ & $43.8$ & $57.6$ & $26.4$ & $68.6$ & $36.6$ & $60.4$ & $25.2$ \\
 & CityWalker~\cite{liu2025citywalker} & $9.8$ & $34.8$ & $42.8$ & $25.6$ & $22.2$ & $0.2$ & $37.6$ & $12.4$ \\
 & InternVLA-DualVLN~\cite{wei2026ground} & $9.6$ & $16.2$ & $28.0$ & $8.4$ & $29.6$ & $20.8$ & $14.6$ & $8.0$ \\
\cmidrule(l){2-10}
 & MIMIC~\cite{he2026learning} & $67.2$ & $84.0$ & $88.2$ & $96.6$ & $33.0$ & $0.0$ & $96.4$ & $69.2$ \\
 & S2E~\cite{he2025seeing} & $22.8$ & $50.2$ & $66.8$ & $45.6$ & $48.4$ & $1.4$ & $83.4$ & $20.4$ \\
 & FlowPilot~\cite{he2026imitation} & $33.8$ & $56.6$ & $57.4$ & $55.4$ & $78.2$ & $8.8$ & $76.8$ & $30.6$ \\
 & OpenPilot~\cite{openpilot} & $11.4$ & $38.6$ & $44.6$ & $82.6$ & $64.6$ & $0.0$ & $99.8$ & $15.6$ \\
\cmidrule(l){2-10}
& \textbf{Avg} & $21.4$ & $38.0$ & $45.7$ & $38.8$ & $41.6$ & $9.1$ & $57.2$ & $21.3$ \\
\midrule
\multirow{10}{*}{\shortstack[l]{\textbf{Minimum}\\\textbf{Pedestrian}\\\textbf{Distance (m)} $\uparrow$}} & ViNT~\cite{shah2023vint} & $1.84$ & -- & $1.14$ & $1.68$ & $1.36$ & -- & $0.80$ & -- \\
 & NoMaD~\cite{sridhar2024nomad} & $1.48$ & $1.23$ & $0.91$ & $1.44$ & $1.07$ & $0.90$ & $0.68$ & -- \\
 & MBRA~\cite{hirose2025learning} & $1.04$ & $0.85$ & $0.76$ & $1.03$ & $0.77$ & $0.78$ & $0.63$ & -- \\
 & CityWalker~\cite{liu2025citywalker} & $1.00$ & $1.06$ & $0.94$ & $0.71$ & $0.89$ & $1.17$ & $0.80$ & $1.72$ \\
 & InternVLA-DualVLN~\cite{wei2026ground} & $1.17$ & $1.26$ & $0.86$ & $1.35$ & $1.88$ & $0.79$ & $0.70$ & -- \\
\cmidrule(l){2-10}
 & MIMIC~\cite{he2026learning} & $0.72$ & $0.74$ & $0.74$ & $0.63$ & $0.98$ & $0.97$ & $0.52$ & -- \\
 & S2E~\cite{he2025seeing} & $0.89$ & $0.77$ & $0.73$ & $0.86$ & $0.91$ & $0.92$ & $0.54$ & $0.91$ \\
 & FlowPilot~\cite{he2026imitation} & $0.87$ & $0.78$ & $0.78$ & $0.76$ & $0.80$ & $0.87$ & $0.62$ & -- \\
 & OpenPilot~\cite{openpilot} & $0.71$ & $0.66$ & $0.62$ & $0.63$ & $0.94$ & $0.95$ & -- & $1.19$ \\
\cmidrule(l){2-10}
& \textbf{Avg} & $1.08$ & $0.92$ & $0.83$ & $1.01$ & $1.07$ & $0.92$ & $0.66$ & $1.27$ \\
\bottomrule
\end{tabular}%
}
\vspace{2mm}
\caption{\textbf{More evaluation results of pedestrian-reactive scenarios}. We evaluate pedestrian collision rate (\%) and average minimum pedestrian distance in the successful
trials. `--' means the model has no successful trials. \textbf{Avg} is the mean across all nine models.} 
\vspace{-4mm}
\label{tab:hand_crafted_full}
\end{table}

More evaluation results of the pedestrian-reactive scenarios are demonstrated in Tab.~\ref{tab:hand_crafted_full}. We additionally compute pedestrian collision rate and average minimum pedestrian distance in the successful trials for evaluating the social compliance of the model. Comparing to Tab.~2  in the main paper, we can see a strong correlation between a high success rate, a low pedestrian collision rate, and a high minimum pedestrian distance. For example, the easiest scenario `Overtaking' has an average success rate of 43.2~\%, a pedestrian collision rate of 9.1~\%, and a minimum pedestrian distance of 0.92~m, while the most challenging scenario ped-crossing has an average success rate of 3.8~\%, a pedestrian collision rate of 57.2~\% and a minimum pedestrian distance of only 0.67~m. This indicates that avoiding collision with pedestrians while maintaining a safe social distance is crucial for the successful completion of pedestrian-reactive scenarios.

\begin{figure}[t]
    \centering
    \includegraphics[width=0.7\textwidth]{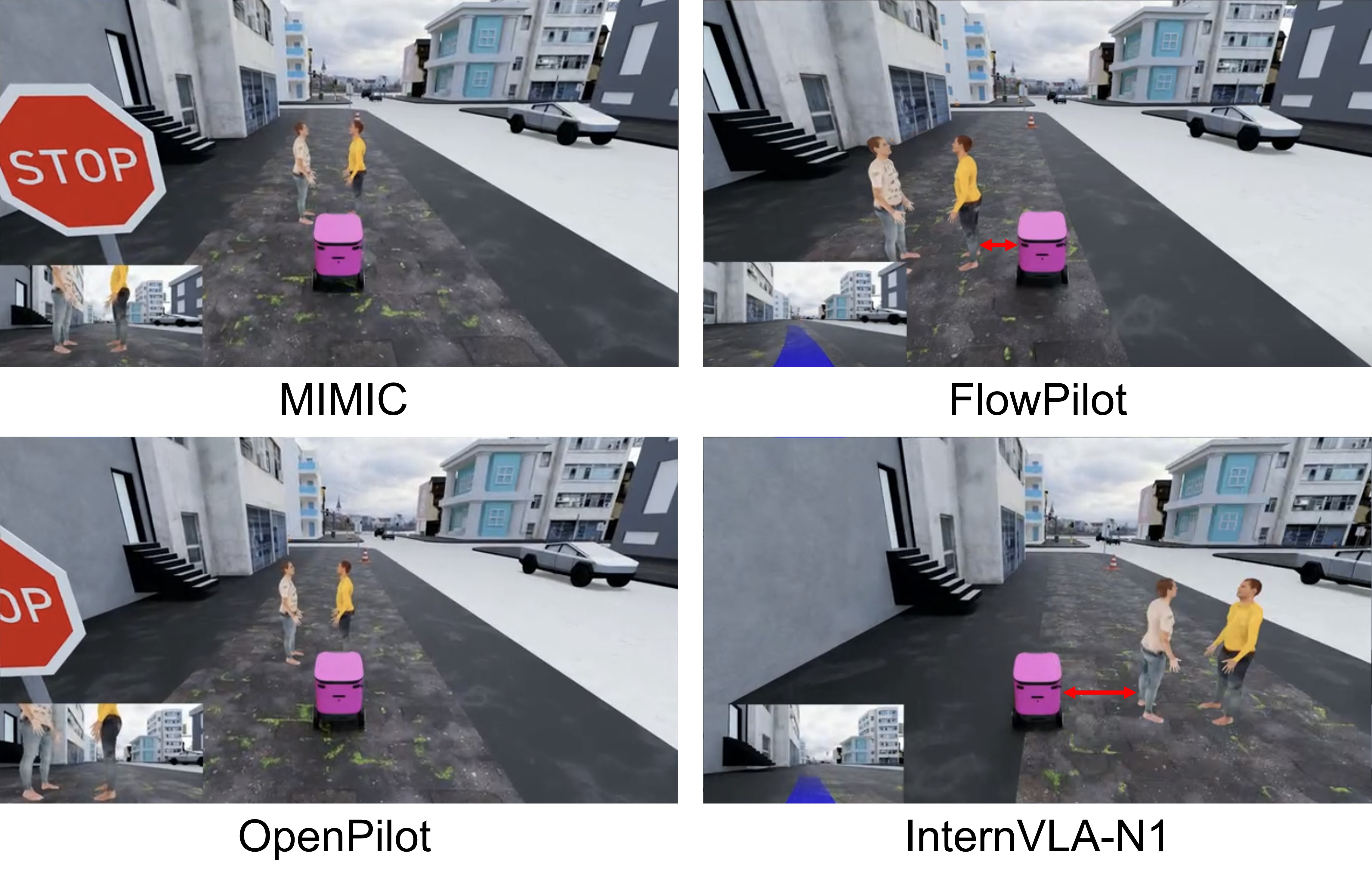}
    \caption{\textbf{Qualitative comparisons of the pedestrian-reactive scenarios.} MIMIC~\cite{he2026learning} and OpenPilot~\cite{openpilot} freeze in front of a conversing pedestrian group. InternVLA-N1~\cite{wei2026ground} leaves a much larger space when passing the  pedestrian group compared to FlowPilot~\cite{he2026imitation}, demonstrating its better social compliance capability than the sidewalk-specific models. We strongly encourage viewing these results (and
other models) as videos on the project page.}
    \label{fig:quali_3}
 
\end{figure}

Moreover, we find that InternVLA-N1~\cite{wei2026ground}, a VLA-based model achieves low pedestrian collision rate and high minimum pedestrian distance in most scenarios despite a relatively low success rate. For example, InternVLA-N1 only achieves an 18.8~\% overall success rate on the challenging  `Frontal Approaching' scenario. However, its pedestrian collision rate of 8.4~\% and  minimum pedestrian distance of 1.35~m  are even better than S2E~\cite{he2025seeing}, the best sidewalk-specific model with a 45.6~\% pedestrian collision rate and a  0.86~m  minimum pedestrian distance. Combining with the results in Tab.~\ref{tab:terrain_only_pg}, we can conclude that while VLA-based models are not good at precise and timely control, they may still be useful in reasoning about the social situations and understanding the pedestrian intents. Therefore, an important future step is combining the low-level sidewalk navigation capabilities of the more light-weight model such as OpenPilot~\cite{openpilot} trained on large-scale  data and the high-level reasoning capabilities from the VLA to better handle the dynamic interaction with the pedestrians.  Some qualitative comparisons of the models are shown in Fig.~\ref{fig:quali_3}.

\section{More Results of  Long-horizon Scenarios}
\label{supp:long_horizon}
\begin{table}[t]
\centering
\definecolor{darkblue}{rgb}{0.75, 0.85, 0.95}  
\definecolor{midblue}{rgb}{0.85, 0.92, 0.98}   
\definecolor{lightblue}{rgb}{0.92, 0.96, 1.0}   

\footnotesize
  \resizebox{0.85\textwidth}{!}{%
  \begin{tabular}{@{}lcccccccccc@{}}
  \toprule
  & \multicolumn{5}{c}{\textbf{Procedurally Generated Scenes}} & \multicolumn{5}{c}{\textbf{Real-World Scanned Scenes}} \\
  \cmidrule(lr){2-6} \cmidrule(lr){7-11}
  \textbf{Method} & \textbf{FC}~$\downarrow$ & \textbf{C}~$\downarrow$ & \textbf{OL}~$\downarrow$ & \textbf{F}~$\downarrow$ &
  $\boldsymbol{V}$~$\uparrow$ & \textbf{FC}~$\downarrow$ & \textbf{C}~$\downarrow$ & \textbf{OL}~$\downarrow$ &
  \textbf{F}~$\downarrow$ & $\boldsymbol{V}$~$\uparrow$ \\
  \midrule
  ViNT~\cite{shah2023vint} & 25.24 & 3.48 & 21.71 & \textbf{0.05} & 1.03 & 24.11 & 7.36 & 16.38 & 0.37 & 1.19 \\
  NoMaD~\cite{sridhar2024nomad} & 8.70 & 2.08 & 6.50 & 0.13 & 1.43 & 5.89 & 2.94 & 2.94 & \textbf{0.00} & \textbf{1.48} \\
  MBRA~\cite{hirose2025learning} & 7.10 & 2.10 & 4.84 & 0.16 & \textbf{1.44} & 5.15 & 2.58 & 2.58 & \textbf{0.00} & 1.46 \\
  CityWalker~\cite{liu2025citywalker} & 11.10 & 1.39 & 7.10 & 2.62 & 0.34 & 10.86 & 4.05 & 1.29 & 5.52 & 0.14 \\
  InternVLA-N1~\cite{wei2026ground} & 12.92 & 2.43 & 10.33 & 0.15 & 1.33 & 9.39 & 4.23 & 4.97 & 0.18 & 1.28 \\
  \midrule
 MIMIC~\cite{he2026learning} & 1.41 & 0.85 & 0.48 & 0.08 & 0.61 & 1.66 & 1.47 & \textbf{0.00} & 0.18 & 0.34 \\
  S2E~\cite{he2025seeing} & 2.36 & 0.69 & 1.53 & 0.14 & 0.85 & 0.92 & 0.74 & 0.18 & \textbf{0.00} & 1.15 \\
  FlowPilot~\cite{he2026imitation} & 1.34 & 0.71 & 0.55 & 0.08 & 1.15 & 0.92 & 0.92 & \textbf{0.00} & \textbf{0.00} & 1.06 \\
  OpenPilot~\cite{openpilot} & \textbf{0.68} & \textbf{0.32} & \textbf{0.24} & 0.12 & 1.19 & \textbf{0.74} & \textbf{0.55} & \textbf{0.00} &
  0.18 & 1.19 \\
  \bottomrule
  \end{tabular}%
  }
\vspace{2mm}

\caption{\textbf{Evaluation results of long-horizon scenarios.}  We report the average failure counts (FC) per 100 meters traveled, including collision (C), out-of-lane (OL), and freezing (F). We also report the average velocity ($V$, m/s).  }
\label{tab:long_horizon}
\end{table}
 Full evaluation results of the long-horizon scenarios are presented in Tab.~\ref{tab:long_horizon}, which corresponds to Fig.~\ref{fig:long-horizon} in the main paper.
\section{Details of Synthetic Training Data Generation}
\label{supp:synthetic}

For the experiments of synthetic training data generation, we collect 500 successful demonstration episodes for both the "Ped-Crossing" and "Gesturing" scenarios in our simulation platform. These demonstrations are generated under the same observation and action representation as the real policy, using ORCA~\cite{van2011reciprocal} to generate collision-free reference trajectories and velocity commands for the robot.
Next, we finetune FlowPilot~\cite{he2026imitation} on the collected synthetic demonstrations. During finetuning, we freeze the visual backbone and only update the lightweight task-adaptation and action-generation modules, mainly the noisy action encoder and action decoder. We use AdamW~\cite{loshchilov2017decoupled} with a learning rate of $5\times10^{-6}$ for all trainable modules. The learning rate is decayed with a cosine annealing schedule for 50 finetuning epochs, with a minimum learning rate of $1\times10^{-6}$. We also apply gradient clipping with a maximum norm of $1.0$ to stabilize optimization. For the real-world experiments, we perform 10 trials for each scenario across 2 distinct locations, and the evaluation protocol strictly matches our simulation
settings.




\end{document}